\documentclass{article}

\usepackage{PRIMEarxiv}

\usepackage[utf8]{inputenc} % allow utf-8 input
\usepackage[T1]{fontenc}    % use 8-bit T1 fonts
\usepackage{hyperref}       % hyperlinks
\usepackage{url}            % simple URL typesetting
\usepackage{booktabs}       % professional-quality tables
\usepackage{amsfonts}       % blackboard math symbols
\usepackage{nicefrac}       % compact symbols for 1/2, etc.
\usepackage{microtype}      % microtypography
\usepackage{lipsum}
\usepackage{fancyhdr}       % header
\usepackage{graphicx}       % graphics
\graphicspath{{media/}}     % organize your images and other figures under media/ folder

\usepackage{enumerate, amsmath, amsthm, amssymb, pifont, csquotes, dashrule, tikz, bbm, booktabs, bm, verbatim, url, float, caption, subcaption}
\usepackage{microtype}
\usepackage{afterpage}
\usepackage[numbers]{natbib}
\usepackage{xspace}

\newcommand{\methodname}{H-DICE}
\newcommand{\expect}{\mathbb{E}}
\newcommand{\mc}[1]{\mathcal{#1}}

\newcommand{\bE}{\mathbb{E}}

\newcommand{\bR}{\mathbb{R}}
\newcommand{\bP}{\mathbb{P}}

\newcommand{\bN}{\mathbb{N}}

\newcommand{\ra}{\rightarrow}

\newcommand\TwoFig[6]{% Image1 Caption1 Label1 Im2 Cap2 Lab2
  \sbox\IBoxA{\includegraphics[width=0.45\textwidth]{#1}}
  \sbox\IBoxB{\includegraphics[width=0.45\textwidth]{#4}}%
  \ifdim\ht\IBoxA>\ht\IBoxB
    \setlength\IHeight{\ht\IBoxB}%
  \else\setlength\IHeight{\ht\IBoxA}\fi
  \begin{figure}[!htb]
  \minipage[t]{0.45\textwidth}\centering
  \includegraphics[height=\IHeight]{#1}
  \caption{#2}\label{#3}
  \endminipage\hfill
  \minipage[t]{0.45\textwidth}\centering
  \includegraphics[height=\IHeight]{#4}
  \caption{#5}\label{#6}
  \endminipage 
  \end{figure}%
}

\title{Hindsight-DICE: Stable Credit Assignment for \newline Deep Reinforcement Learning
}

\author{
  Akash Velu$^*$ \hspace{10pt} 
  Skanda Vaidyanath$^*$ \hspace{10pt}
  Dilip Arumugam \\
  Department of Computer Science \\Stanford University \\
  \texttt{\{avelu,svaidyan,dilip\}@stanford.edu} \\
}

\begin{document}

\def\thefootnote{*}\footnotetext{Equal contribution.}
\renewcommand*{\thefootnote}{\arabic{footnote}}
\setcounter{footnote}{0}

\maketitle

\begin{abstract}
    Oftentimes, environments for sequential decision-making problems can be quite sparse in the provision of evaluative feedback to guide reinforcement-learning agents. In the extreme case, long trajectories of behavior are merely punctuated with a single terminal feedback signal, leading to a significant temporal delay between the observation of a non-trivial reward and the individual steps of behavior culpable for achieving said reward. Coping with such a credit assignment challenge is one of the hallmark characteristics of reinforcement learning. While prior work has introduced the concept of hindsight policies \cite{NEURIPS2019_195f1538} to develop a theoretically motivated method for reweighting on-policy data by impact on achieving the observed trajectory return, we show that these methods experience instabilities which lead to inefficient learning in complex environments. In this work, we adapt existing importance-sampling ratio estimation techniques for off-policy evaluation to drastically improve the stability and efficiency of these so-called hindsight policy methods. Our hindsight distribution correction facilitates stable, efficient learning across a broad range of environments where credit assignment plagues baseline methods.

    % Oftentimes, environments for sequential decision-making problems can be quite sparse in the provision of evaluative feedback to guide reinforcement-learning agents. In the extreme case, long trajectories of behavior are merely punctuated with a single terminal feedback signal, engendering a significant temporal delay between the observation of non-trivial reward and the individual steps of behavior culpable for eliciting such feedback. Coping with such a credit assignment challenge is one of the hallmark characteristics of reinforcement learning and, in this work, we capitalize on existing importance-sampling ratio estimation techniques for off-policy evaluation to drastically improve the handling of credit assignment with policy-gradient methods. While the use of so-called hindsight policies offers a principled mechanism for reweighting on-policy data by saliency to the observed trajectory return, naively applying importance sampling results in unstable or excessively lagged learning. In contrast, our hindsight distribution correction facilitates stable, efficient learning across a broad range of environments where credit assignment plagues baseline methods.

\end{abstract}

% keywords can be removed
% \keywords{First keyword \and Second keyword \and More}

\section{Introduction}

Reinforcement learning is the classic paradigm for addressing sequential decision-making problems~\citep{sutton1998introduction}. Naturally, while inheriting the fundamental challenge of generalization across novel states and actions from supervised learning, general-purpose reinforcement-learning agents must also contend with the additional challenges of exploration and credit assignment. While much initial progress in the field was driven largely by innovative machinery for tackling credit assignment~\citep{sutton1984temporal,sutton1988learning,singh1996reinforcement} alongside simple exploration heuristics ($\varepsilon$-greedy exploration, for example), recent years have seen a reversal with the bulk of attention focused on a broad array of exploration methods (spanning additional heuristics as well as more principled approaches) \cite{tang2017exploration, pathak2017curiositydriven, ecoffet2021goexplore}, and relatively little consideration given to issues of credit assignment. This lack of interest in solution concepts, however, has not stopped the proliferation of reinforcement learning into novel application areas characterized by long problem horizons and sparse reward signals; indeed, the current reinforcement learning from human feedback (RLHF) paradigm~\citep{huggingface2022rlhf} is now a widely popularized example of an environment that operates in perhaps the harshest setting where a single feedback signal is only obtained after the completion of a long trajectory. \footnote{Our implementation is available at \hyperlink{https://github.com/skandavaidyanath/credit-assignment}{https://github.com/skandavaidyanath/credit-assignment}.}

% \snote{a note on RLHF, goal reaching, language conditioned}
% Tremendous strides have been made in recent years by leveraging human feedback as a learning signal for reinforcement-learning agents, with particularly impressive results for fine-tuning large language models~\citep{ziegler2019fine}. One of the inescapable realities of human-in-the-loop reinforcement learning, regardless of whether that feedback is modeled as rewards~\citep{Isbell2001ASR,thomaz2006reinforcement,knox2008tamer,pilarski2011online,knox2012learning,warnell2018deep}; advantages~\citep{macglashan2017interactive,arumugam2019deep}; or preferences~\citep{akrour2011preference,wilson2012bayesian,akrour2012april,furnkranz2012preference,akrour2014programming,wirth2016model,el2016score,christiano2017deep,ibarz2018reward,leike2018scalable}, is that a human evaluator can only be queried for feedback a limited number of times before feeling overburdened, unlike the automated feedback mechanisms used widely throughout the classic literature~\citep{bellemare2013arcade}. Indeed, the current reinforcement learning from human feedback (RLHF) paradigm~\citep{huggingface2022rlhf} is now a widely popularized example of an environment that operates in perhaps the harshest setting where a single feedback signal is only obtained after the completion of a long trajectory. 

This combination of long horizons with delayed rewards most directly impacts an agent's ability to efficiently perform credit assignment~\citep{minsky1961steps,sutton1984temporal,sutton1988learning} and thereby worsens the overall sample complexity needed to synthesize an optimal policy. Therefore, while a reinforcement-learning agent must also contend with the challenges of generalization and exploration, it is perhaps the credit assignment challenge that poses the greatest impediment to sample-efficient reinforcement learning. While theoretical results encapsulating settings like that of the RLHF paradigm are nascent~\citep{chatterji2021theory,pacchiano2023dueling}, practical and demonstrably-scalable mechanisms for efficient credit assignment are missing from the literature.

With this reality in mind, our work extends the relatively recent solution concept of hindsight policies as introduced by~\citet{NEURIPS2019_195f1538}, which allow an agent to reason counterfactually about past decisions at particular states in light of observed future outcomes. While the natural course of action for algorithm design based on the provision of such a hindsight policy is to reweight the on-policy data collected for agent updates, we highlight a key issue where the resulting importance-sampling ratio is highly unstable, compromising the quality of learning. Even when ratio clipping is applied, as one often does in off-policy policy-gradient methods~\citep{schulman2017proximal}, such a heuristic lends stability at the heavy cost of incredibly dampened learning speed.

To remedy this instability, we turn to prior work tackling the orthogonal problem of off-policy policy evaluation (OPE)~\citep{precup2000eligibility,jiang2016doubly,thomas2016data} wherein the ratio between state visitation probabilities of a behavior policy and target policy has emerged over recent years as a key quantity of interest for yielding principled, practical OPE algorithms~\citep{hallak2017consistent,liu2018breaking,gelada2019off,liu2019off,liu2020understanding,nachum2019dualdice,uehara2020minimax,zhang2020gendice,zhang2020gradientdice,dai2020coindice} whose variance does not suffer exponentially in the problem horizon, as it would with a naive application of importance sampling. Inspired by the potential for technical progress in this separate arena to be recycled and operationalized towards achieving more efficient credit assignment techniques, we select one such approach from the OPE literature, Dual stationary DIstribution Correction Estimation (DualDICE)~\citep{nachum2019dualdice}, and adapt the importance ratio approximation technique for hindsight policies. Our resulting \textbf{H}indsight \textbf{DI}stribution \textbf{C}orrection \textbf{E}stimation \textbf{(\methodname{})} approach empowers policy-gradient methods~\citep{schulman2017proximal} to empirically avoid the same catastrophic instability that arises by naively computing the importance ratio between the current and hindsight policies. We empirically demonstrate over three different and challenging benchmarks tasks that our method outperforms baseline methods, which either do not incorporate any form of credit assignment or do so in a naive fashion, by a large margin both in terms of the achieved final rewards and the speed of convergence. 

The paper proceeds as follows: in Section \ref{sec:back} we detail our problem formulation as well as background information on credit assignment with hindsight policies. In Section \ref{sec:approach}, we outline our approach before presenting a detailed empirical evaluation and discussion of results in Section \ref{sec:exps}. We conclude with an overview of related work in Section \ref{sec:related} and future work in Section \ref{sec:conc}.

\section{Preliminaries \& Background}
\label{sec:back}

In this section, we begin by specifying our problem formulation before providing brief background information on the credit assignment problem broadly as well as the specific use of hindsight policies to address the challenge.

For any natural number $N \in \bN$, we denote the index set as $[N] \triangleq \{1,2,\ldots,N\}$. For any set $\mc{X}$, we denote the set of all probability distributions with support on $\mc{X}$ as $\Delta(\mc{X})$. For another arbitrary set $\mc{Y}$, we denote the class of all functions mapping from $\mc{X}$ to $\mc{Y}$ as $\{\mc{X} \ra \mc{Y}\}$.

\subsection{Problem Formulation}

We formulate a sequential decision-making problem as an infinite-horizon, discounted Markov Decision Process (MDP)~\citep{bellman1957markovian,Puterman94} defined by $\mc{M} = \langle \mc{S}, \mc{A}, \mc{R}, \mc{T}, \mu, \gamma \rangle$. Here $\mc{S}$ denotes a set of states, $\mc{A}$ is a set of actions, $\mc{R}:\mc{S} \times \mc{A} \ra \bR$ is a deterministic reward function providing evaluative feedback signals to the agent, $\mc{T}:\mc{S} \times \mc{A} \ra \Delta(\mc{S})$ is a transition function prescribing distributions over next states, $\mu \in \Delta(\mc{S})$ is an initial state distribution, and $\gamma \in [0,1)$ is the discount factor communicating a preference for near-term versus long-term rewards. Beginning with an initial state $s_0 \sim \mu$, for each timestep $t \in \bN$, the agent observes the current state $s_t \in \mc{S}$, selects action $a_t \sim \pi(\cdot \mid s_t) \in \mc{A}$, enjoys a reward $r_t = \mc{R}(s_t,a_t)$, and transitions to the next state $s_{t+1} \sim \mc{T}(\cdot \mid s_t, a_t) \in \mc{S}$. 

Our reinforcement-learning approach is grounded in policy search~\citep{williams1992simple,sutton1999policy,konda1999actor,mnih2016asynchronous,schulman2017proximal} wherein the agent's behavior is represented by a stationary, stochastic policy $\pi_\theta:\mc{S} \ra \Delta(\mc{A})$, parameterized by $\theta \in \Theta \subset \bR^d$, which encodes a pattern of behavior mapping individual states to distributions over possible actions. The overall performance of $\pi_\theta$ in any MDP $\mc{M}$ when starting at state $s \in \mc{S}$ and taking action $a \in \mc{A}$ is assessed by its associated action-value function $Q^{\pi_\theta}(s,a) = \bE\left[\sum\limits_{t=0}^\infty \gamma^t \mc{R}(s_{t},a_{t}) \bigm| s_0 = s, a_0 = a\right]$, where the expectation integrates over randomness in the action selections and transition dynamics. Taking the corresponding value function as $V^{\pi_\theta}(s) = \bE_{a \sim \pi_\theta(\cdot \mid s)}\left[Q^{\pi_\theta}(s,a)\right]$ and letting $\Pi_\Theta \triangleq \{\pi_\theta \mid \theta \in \Theta\} \subset \{\mc{S} \ra \Delta(\mc{A})\}$ denote the policy class parameterized by $\Theta$ (a subset of all stochastic policies), we define the optimal policy $\pi^\star$ as achieving supremal value with respect to this class $V^\star(s) = \sup\limits_{\pi \in \Pi_\Theta} V^\pi(s) = \max\limits_{a^\star \in \mc{A}} Q^\star(s,a^\star)$ and $Q^\star(s,a) = \sup\limits_{\pi \in \Pi_\Theta} Q^\pi(s,a)$ for all $s \in \mc{S}$ and $a \in \mc{A}$. As the agent interacts with the environment over the course of $K \in \bN$ episodes, we let $\tau_k = (s^{(k)}_1, a^{(k)}_1, r^{(k)}_1, \ldots) \sim \rho^{\pi_{\theta_k}}$ be the random variable denoting the trajectory experienced by the agent in the $k$th episode sampled from the induced trajectory distribution $\rho^{\pi_{\theta_k}}$ of policy $\pi_{\theta_k}$, for any $k \in [K]$. Alternatively, one may also examine the visitation of a policy $\pi_\theta$ through its discounted, stationary state distribution: $d^{\pi_\theta}(s) = (1-\gamma) \sum\limits_{t=0}^\infty \gamma^t \bP^{\pi}(s_t = s),$ where $\bP^{\pi}(s_t = \cdot) \in \Delta(\mc{S})$ denotes the distribution over states visited by $\pi_\theta$ at each timestep $t$. Finally, in alignment with prior work~\citep{10.5555/3305381.3305428,NEURIPS2019_195f1538}, we use $Z(\tau)$ to denote the random variable representing the return of a trajectory $\tau$.

% \snote{We don't actually use a lot of the notation here. Also we're using just $z$ for return and not $Z(\tau)$}
% \snote{Actually on second thought, maybe the only thing we really don't need is the trajectory random variable and the index set of natural numbers...?}
% \anote{We'd need to define advantage, trajectory RV, index set, discounted state/action distributions I think.}

%\dnote{Just some initial notation that we should adjust as needed.}

\begin{figure*}[t!]
\centering
\begin{subfigure}{.5\textwidth}
  \centering
  \includegraphics[width=0.75\linewidth]{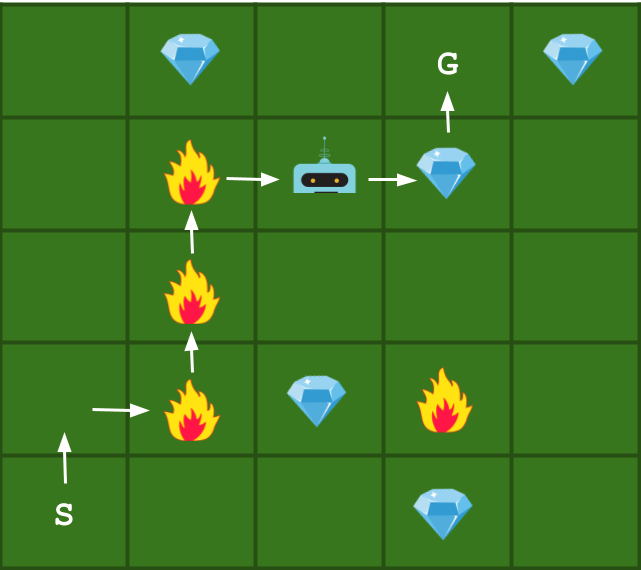}
  \label{fig:intro_sub1}
\end{subfigure}%
\begin{subfigure}{.5\textwidth}
  \centering
  \includegraphics[width=1.0 \linewidth]{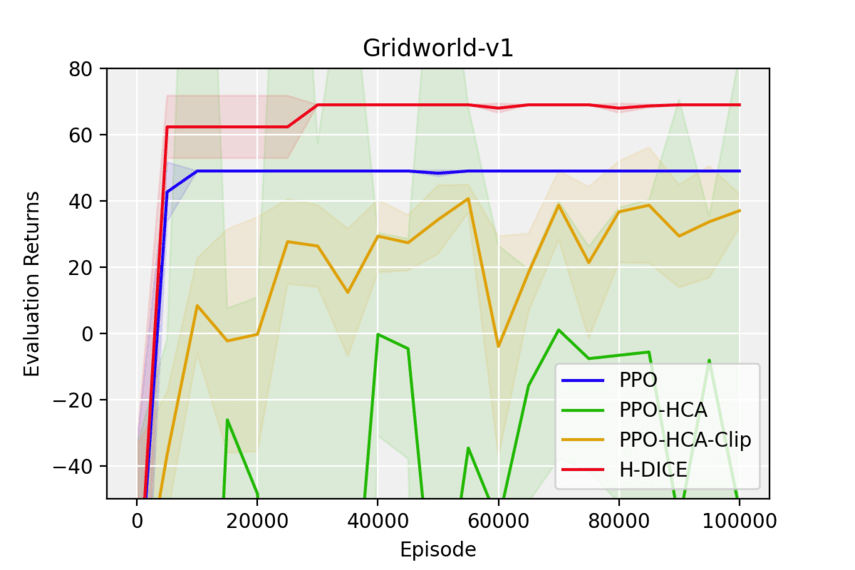}
  \label{fig:intro_sub2}
\end{subfigure}
\caption{\emph{Left:} GridWorld-v1 environment, in which the agent starts at square \emph{S} and must collect diamonds to obtain positive rewards, and avoid fire which result in negative rewards. The reward signal is delayed so that the agent receives the sum of all rewards upon episode termination, which occurs either when the agent transitions into the \emph{G} square or the episode times out. Solid arrows denote an example trajectory in which the agent steps into fire three times and  collects one diamond before reaching the $G$ square. The robot indicates the position of the agent at a point in this trajectory, referenced in Section \ref{sec:interpretability}. \emph{Right:} Learning curves of our proposed method, \methodname{}, and baseline methods (PPO, PPO-HCA, and PPO-HCA-Clip) on Gridworld-v1. \methodname{} achieves the maximum reward quickly and stably compared to other methods. Vanilla PPO is unable to disentangle the good and bad actions even in this simple environment since it does not explicitly model credit assignment and this leads to a suboptimal policy. While using unclipped HCA ratios to compute advantages (PPO-HCA) results in highly unstable performance, clipping the ratios (PPO-HCA-Clip) improves results slightly.}
%\dnote{A reader might be left wondering how/why PPO converges to a sub-optimal return value.}
\label{fig:intro}
\end{figure*}

\subsection{A Motivating Example for Credit Assignment}
 Consider the gridworld setting depicted in Figure \ref{fig:intro}, wherein an agent accrues positive rewards for collecting diamonds and negative rewards for moving through fire. Importantly, however, the agent only receives a nonzero reward signal -- which is the sum of the rewards accrued in each individual transition -- at the end of a trajectory when it enters the $G$ square and no receives rewards for the intermediary steps. 

In the example trajectory illustrated in Figure \ref{fig:intro}, the agent transitions through three fire squares before acquiring a diamond and reaching $G$, upon which it receives a large negative reward (due to the three steps into the fire). A typical policy gradient method would place ``blame'' for this poor outcome on \emph{each} action in the trajectory and would correspondingly reduce the probability that these actions are taken by the policy. In fact, due to a discount factor $\gamma < 1$, the algorithm would assign greater blame to actions which are temporally closer to the final return, resulting in a larger decrease in the probability of the action which picked up the diamond compared to the actions which put the agent in fire; this, however, is undesirable. Despite the final negative return, picking up the diamond is the optimal action in the corresponding square. Ideally, this action should be made more likely, while the actions moving the agent into fire should be reduced in likelihood, as they are culpable for the negative return. 

Credit assignment aims to perform exactly this type of reasoning; not all actions are equally responsible for a future outcome, and determining which actions are responsible in yielding a particular end result -- which is the goal of credit assignment -- can enable more efficient policy learning and improved policy performance. 

% \anote{The ending of this paragraph feels a bit weak, but not sure yet as to how we can improve it.}

\subsection{Hindsight Credit Assignment}
\label{sec:hca}
 More generally, the credit assignment problem revolves around the fixed behavior policy of an agent and asks \textit{what is the impact of being in a particular state and executing a particular action on observed future outcomes?}~\citep{NEURIPS2019_195f1538,arumugam2021information}. Typically, the future outcome of interest is the full return $Z(\tau)$ associated with a complete trajectory $\tau$, although one could also opt to examine intermediate returns observed from any incident states or state-action pairs. 
% Unfortunately, various problem-specific axes and dimensions emerge in the reinforcement-learning setting that can exacerbate the credit assignment problem and make it exceedingly difficult to obtain the answer to this fundamental question. 
Unfortunately, various aspects of the reinforcement-learning setting exacerbate the credit assignment problem and make it exceedingly difficult to accurately answer this fundamental question in a straightforward manner. 
Long horizons enable feedback structures where nontrivial signals can emerge after a significant delay from the culpable step of behavior. Environments whose underlying transition structure is highly stochastic induce large variance in the resulting trajectories sampled from any policy, making it difficult to obtain accurate estimates of the overall return needed to assign credit or blame. In the context of policy search, initial policies whose actions carry little information about the incident trajectory returns can make initial thrust toward efficiently allocating credit to observed experiences challenging~\citep{arumugam2021information}. Moreover, credit assignment must often be adjudicated in the presence of a single sampled trajectory but, ideally, should factor in this stochasticity to give appropriate credit to those steps of behavior that could have occurred under a different random rollout~\citep{van2021expected}.

While classic approaches to addressing credit assignment would naively distribute credit based on temporal recency to interesting events or outcomes~\citep{klopf1972brain,sutton1984temporal,singh1996reinforcement}, a more recent and cogent mechanism has emerged through the use of a \textit{hindsight policy}~\citep{NEURIPS2019_195f1538}. For the purposes of this work, a hindsight policy $h^{\pi_\theta}_\omega: \mc{S} \times \bR \ra \Delta(\mc{A})$ induced by a behavior policy $\pi_\theta$ and parameterized by $\omega \in \Omega \subset \bR^m$ is a return-conditioned policy that yields the distribution over actions taken by policy $\pi_\theta$ from a given state and in light of an observed trajectory return. Inspired by the hindsight experience replay technique~\citep{andrychowicz2017hindsight} for goal-based reinforcement learning, a hindsight policy allows an agent to consider the full space of possible actions that could have occurred in hindsight based on the final trajectory return. This effect becomes apparent from the resulting identity for the advantage function~\citep{baird1993advantage} computed by policy-gradient methods~\citep{mnih2016asynchronous,schulman2017proximal,schulman2018highdimensional}: 
\begin{align}
    A^{\pi_\theta}(s,a) &= Q^{\pi_\theta}(s,a) - V^{\pi_\theta}(s) = \mathbb{E}\left[\left(1 - \frac{\pi_\theta(a \mid s)}{h^{\pi_\theta}_{\omega}(a \mid s, Z(\tau))}\right) Z(\tau)\right].
    \label{eq:hca_advantage}
\end{align}

% \vspace{1mm}
% \begin{align}
%     A^{\pi_\theta}(s,a) &= Q^{\pi_\theta}(s,a) - V^{\pi_\theta}(s) \nonumber  \\
%     &= \mathbb{E}\left[\left(1 - \frac{\pi_\theta(a \mid s)}{h^{\pi_\theta}_{\omega}(a \mid s, Z(\tau))}\right) Z(\tau)\right].
%     \label{eq:hca_advantage}
% \end{align}

If, in hindsight, observing a return $Z(\tau)$ makes a particular action less likely, then the \emph{hindsight ratio} $\frac{\pi_\theta(a \mid s)}{h^{\pi_\theta}_{\omega}(a \mid s, Z(\tau))} > 1$ and detracts from the advantage of the action; in contrast, an action that makes the return $Z(\tau)$ more plausible in hindsight results in an increased advantage signal as $\frac{\pi_\theta(a \mid s)}{h^{\pi_\theta}_{\omega}(a \mid s, Z(\tau))} < 1$. In the next section, we discuss the practical challenges that emerge when trying to estimate this hindsight ratio so as to integrate this credit assignment information within a policy-gradient method.

\section{Approach}
\label{sec:approach}

We begin by examining the natural, straightforward approach to hindsight ratio estimation and highlight stability issues that arise from using such an estimator. We then proceed to introduce one alternative avenue for hindsight ratio estimation inspired by prior work in OPE. Finally, we conclude by introducing our \methodname{} approach, which leverages this ratio estimator for demonstrably stable credit assignment.

\subsection{Direct Density Ratio Instability} \label{sec: instability}
As detailed in Section \ref{sec:hca}, the HCA framework derives an expression for advantage function $A^{\pi_\theta}(s, a)$ of a policy $\pi_\theta$, which, through the return-conditioned hindsight function $h^{\pi_\theta}$, explicitly assigns the credit each action receives for achieving a particular return from the current state. Given on-policy trajectory data $\mc{D}^{\pi_\theta}$, a straightforward way to compute this advantage would be to first fit a hindsight function $h^{\pi_\theta}_\omega$ to the data via supervised learning (where $\omega$ are trainable parameters). 
The hindsight ratio is then calculated via $\frac{\pi_\theta(a \mid s)}{h^{\pi_\theta}_{\omega}(a \mid s, Z(\tau))}$ for each state-action pair $(s, a)$ in a trajectory $\tau$, and $A^{\pi}(s, a)$  can subsequently be computed using expression \ref{eq:hca_advantage}, as done in the approach of \citet{NEURIPS2019_195f1538}.

However, direct estimation of density ratios can be highly unstable and, as shown in the right of Figure \ref{fig:intro} and Section \ref{sec:interpretability}, the hindsight ratio is no exception. Using Proximal Policy Optimization (PPO)~\citep{schulman2017proximal} as a base policy-gradient algorithm, we empirically observe that values the ratio takes through the course of training vary greatly, hindering policy learning. PPO-HCA, a method which directly estimates the hindsight ratio, performs poorly and exhibits high variance across seeds. Clipping the ratios to be between zero and one, a common way to stabilize density ratios~\citep{schulman2017proximal}, improves the stability of learning but still results in suboptimal performance, as seen by the results of PPO-HCA-Clip. 
%More broadly, such a clipping approach introduces a hyperparameter that must be tuned on a per-problem basis in order to maximize the benefits of HCA while retaining stability throughout the learning process. 
Thus, the question remains: how can we compute the hindsight ratio in a \emph{stable} manner?

% \snote{Its unclear what the reader should look for in the figure here because we haven't really explained what each line in the plot refers to}

\subsection{Distribution Correction Estimation}
In this work, we offer one candidate answer to the above question by taking inspiration from the off-policy policy evaluation (OPE) literature, where recent methods have been developed to accurately and soundly compute the importance sampling ratio between two state-action distributions. Specifically, we leverage the following observation from \cite{nachum2019dualdice} which serves as a core building block for their OPE algorithm: defining $\mathcal{C} \triangleq [0, C]$ for a bounded constant $C \in \bR_{\geq 0}$ and a function $x: \mc{S} \times \mc{A} \ra \mc{C}$, the solution $x^*(s, a)$ of the following optimization problem
\begin{align}
    \min_{x: S \times A \rightarrow \mathcal{C}} \frac{1}{2}\expect_{(s, a) \sim d^{\mc{D}}}[x(s, a)^2] - \expect_{(s, a) \sim d^{\pi}}\left[ x(s, a) \right]
    \label{eq:dice}
\end{align}
is given by $x^*(s, a) = \frac{d^{\pi}(s, a)}{d^{\mc{D}}(s, a)}$.
In the context of their work, $d^{\pi}(s, a)$ is the discounted state-action visitation distribution of a target policy $\pi$ of interest while $d^{\mc{D}}(s, a)$ is free to be an arbitrary empirical distribution over state-action pairs representing a static, offline dataset $\mc{D}$ (which potentially emerges from sampling a single behavior policy, multiple policies, or some other sampling oracle). 
In the OPE setting, the second term of Equation \ref{eq:dice} is problematic as collecting on-policy rollouts of $\pi$ is not possible. Consequently, the DualDICE method \cite{nachum2019dualdice} goes on to extend on this optimization and remedy the issue. In contrast, the on-policy nature of the credit assignment problem, whose core question is always asked with respect to a current behavior policy of interest, makes Equation \ref{eq:dice} a sufficient and viable approach for hindsight ratio estimation. We leave to future work the question of how alternative techniques for density-ratio estimation~\citep{choi2022density} might be applied towards achieving stability and improved agent performance in the context of HCA.

\subsection{Hindsight Distribution Correction Estimation} \label{sec: HDICE}
Our \methodname{} method for stable computation of the hindsight ratio $\frac{\pi_\theta(a \mid s)}{h^{\pi_\theta}_{\omega}(a \mid s, Z(\tau))}$ adapts the observation from the previous section to our setting by instead focusing on the following optimization:
\begin{equation} 
    \min_{\phi: S \times A \times Z \rightarrow \mathcal{C}} \frac{1}{2}\expect_{(s, a, z) \sim \mc{D}^{{h}^{\pi_\theta}}}[\phi(s, a, z)^2] - 
     \expect_{\substack{(s, a) \sim d^{\pi_\theta}\\ z \sim \psi(z)}}\left[ \phi(s, a, z) \right]
\label{eq:h-dice-opt} 
\end{equation}

where $\mc{D}^{{h}^{\pi_\theta}}$ denotes an empirical state-action-return distribution formed by drawing states under the $d^{\pi_\theta}$ visitation of $\pi_\theta$, returns of $\pi_\theta$ observed from state $s$ according to $\chi^{\pi_\theta}(\cdot |s)$, and actions from the hindsight policy $h^{\pi_\theta}_\omega(\cdot \mid s, z)$. 
% Meanwhile, $\psi(z)$ is an arbitrary distribution over the returns that, in our experiments, we model as a Dirac delta distribution centered on $z=0$, which results in simplified hindsight ratio calculation.
Meanwhile, $\psi(z)$ is an arbitrary distribution over the returns that, in our experiments, we model as a uniform distribution over all possible returns in the environment, which results in simplified hindsight ratio calculation. In the main results presented in Section \ref{sec:exps}, we set the range threshold $C \in \bR_{\geq 0}$ for $\phi$ to be 1; however, as shown in Appendix \ref{app:c-ablation}, the choice of $C$ is largely inconsequential and numerous values result in similar performance.

While Appendix \ref{app:math} elaborates in more detail, we briefly mention here that the solution to this optimization, $\phi^{\star}$, is given by: $$\phi^{\star}(s, a, z) = \frac{\pi_\theta(a | s)}{\chi^{\pi_\theta}(z|s)h^{\pi_\theta}_\omega(a|s, z)}.$$
% \begin{align*}
%     \phi^{\star}(s, a, z) &= \frac{d^{\pi}(s, a) \psi(z=0)}{d^{h^{\pi}}(s, a, z)} \\ 
%     &= \frac{d^{\pi}(s)\pi(a | s)\psi(z=0)}{d^{h^{\pi}}(s)\chi^\pi(z|s)h^{\pi}(a|s, z)} \\ 
%     &= \frac{\pi(a | s)}{\chi^\pi(z|s)h^{\pi}(a|s, z)}
% \end{align*}
 
% The final simplification can be made because $\psi(z=0) = 1$ as we have chosen above, and empirically we pass in 0 returns to the model when we evaluate this expectation.
% $d^{\pi}(s) = d^{h^{\pi}}(s)$ follows from the fact that $h^{\pi}$ is the hindsight policy learned from the state-visitation distribution of $\pi$. 
% Empirically, we use the same data that we collected from the policy to train the hindsight distribution as well and we update them with the same schedule.
 
This allows us to arrive at the following expression for the hindsight ratio of interest:

\begin{equation} 
    \frac{\pi_\theta(a | s)}{h^{\pi_\theta}_\omega(a | s, z)} = \phi^{\star}(s, a, z)\chi^{\pi_\theta}(z|s)
    \label{eq:H-dice-final}
\end{equation}

% We have included a more detailed derivation of the objective function, justifying the choices made and show that minimizing it leads to this solution in Appendix section \ref{app:math}.

Importantly, the right hand side of Equation \ref{eq:H-dice-final} can be estimated from on-policy data $\mc{D}^{\pi_\theta}$ collected under $\pi_\theta$. 
This involves learning the following three models, which we instantiate as neural networks:
\begin{enumerate}
    \item \textbf{Return predictor $\chi^{\pi_\theta}_\eta$}: We learn the distribution over returns incurred by policy $\pi_\theta$ given a state $s \in \mc{S}$ via supervised learning using $\mc{D}^{\pi_\theta}$. Note that our approach can be seen as a Monte-Carlo estimation of the return distribution and $\chi^{\pi_\theta}_\eta$ could also be learned via distributional reinforcement learning \cite{10.5555/3305381.3305428}, an avenue we leave to future work.
    \item \textbf{Hindsight policy $h^{\pi_\theta}_\omega$}: Just as in \citet{NEURIPS2019_195f1538}, we learn a hindsight policy via supervised learning using on-policy data $\mc{D}^{\pi_\theta}$.
    \item \textbf{Hindsight DICE model $\phi_\nu$}: Lastly, our estimate of $\phi^\star$ in Equation \ref{eq:H-dice-final} is given by $\phi_\nu$ optimized via Equation \ref{eq:h-dice-opt}. Naturally, this requires $\mc{D}^{\pi_\theta}$ along with $\chi^{\pi_\theta}_\eta$ and $h^{\pi_\theta}_\omega$.
\end{enumerate}

As all the aforementioned models used to compute hindsight ratios are (either implicitly or explicitly) conditioned on the current policy, $\pi_\theta$, we must reset the weights of the corresponding function approximators after each policy update and before collecting fresh data from the latest policy. In our experiments, since these models are implemented as lightweight, fully-connected networks with few layers, we find that this does not adversely affect wall-clock time or the computational efficiency of our algorithm. 
% \anote{Update this paragraph to allude to the OP ablation we do.}

Like DualDICE, our approach for estimating the hindsight ratio inherits an avoidance of explicitly computing importance-sampling weights, thereby alleviating the instability that arises from direct density ratio estimation observed in Section \ref{sec: instability}. The full algorithm for policy learning using \methodname{} is detailed in Algorithm \ref{Alg: H-DICE}. Note that although H-DICE and all baseline methods use PPO \cite{schulman2017proximal} as the base policy learning algorithm, our approach is agnostic to this choice and compatible with any policy-gradient algorithm. In contrast to many standard PPO implementations which learn a value-function alongside a policy to compute advantages,  \methodname{} does not learn a value-function to compute advantages as a consequence of Equation \ref{eq:hca_advantage}.

\begin{algorithm}[tb] 
   \caption{Hindsight Distribution Correction Estimation}
    \label{Alg: H-DICE}
\begin{algorithmic}
    \STATE{Initialize $\pi_{\theta_1}$} \vspace{0.5mm}
   \FOR{iteration $i=1 \dots N$} \vspace{0.5mm}
   \STATE{Initialize $h^{\pi_{\theta_i}}_\omega$, $\phi_\nu$, and $\chi^{\pi_{\theta_i}}_\eta$} \vspace{0.5mm}
   \STATE{Collect dataset $\mc{D}^{\pi_{\theta_i}}$ of on-policy trajectories with $\pi_{\theta_i}$} \vspace{0.5mm}
   \STATE{Train $\chi^{\pi_{\theta_i}}_\eta$ and $h^{\pi_{\theta_i}}_\omega$ by supervised learning with $\mc{D}^{\pi_{\theta_i}}$} \vspace{0.5mm}
   % \STATE{Train hindsight policy $h^{\pi}$ via supervised learning with $D_i$ for $E_{\text{hindsight}}$ epochs.}
   % \vspace{0.5mm}
   % \FOR{each state, return pair $s_t, z_t$ in $D_i$}
   % \STATE{Compute $a_t^h \sim h^{\pi}(a | s, z)$ and store in $D^{\text{hindsight}}_i$}
   % \ENDFOR
   % \vspace{0.5mm}
   \STATE{Train \methodname{} $\phi_\nu$ by Equation \ref{eq:h-dice-opt}}
    \vspace{0.5mm}
    \STATE{Obtain $\theta_{i+1}$ via PPO update on $\pi_{\theta_i}$ with advantages in Equation \ref{eq:hca_advantage}}
    \vspace{0.5mm}
   % \STATE{Compute Advantages using Equation \ref{eq:hca_advantage}}.
   % \vspace{0.5mm}
   % \STATE{Perform a PPO update on $\pi$.  }
   % \vspace{0.5mm}
   % \STATE{Reset the parameters of $h^{\pi_{\theta_i}}_\omega, \phi^{\pi_{\theta_i}}_\nu, \chi^{\pi_{\theta_i}}_\eta$.}
   \ENDFOR
\end{algorithmic}
\end{algorithm}

\section{Experiments}
\label{sec:exps}

In this section, we evaluate \methodname{} on complex discrete and continuous control tasks and compare it with baselines.
We also closely investigate the learned hindsight ratios and explore varying update schedules for the hindsight, return and DICE models (referred to jointly as ``auxiliary models'').
For the sake of brevity, we defer additional experiments on the impact of the constant $C$, using different distributions for $\psi$, and using the \methodname{} advantage estimator on dense reward settings to Appendix Sections \ref{app:c-ablation}, \ref{app:psi-ablation} and \ref{app:dense} respectively. We begin by providing the details of our evaluation domains and chosen baselines.

% We seek to answer the following through our main experiments: \vspace{-7pt}
% \begin{enumerate}
%     \item Does performing credit assignment via H-DICE result in better final performance and learning efficiency compared to methods which do not do credit assignment?
%     \item How does H-DICE compare to simpler methods of stabilizing the hindsight ratio in terms of learning stability and final performance?
%     % \item How does H-DICE perform in discrete and continuous action settings?
%     \item What is the impact of episode length on the importance of credit-assignment?
% \end{enumerate}

% \vspace{-10pt}
\subsection{Domains} \label{Sec: domains}
We evaluate methods on the following domains: 

\textbf{GridWorld:} The GridWorld environments, shown in Figures \ref{fig:intro} and \ref{fig:Gridworld-v2-img}, are environments we develop which pose a difficult credit-assignment challenge. In each square, the agent can take actions representing the cardinal directions.
%\dnote{TODO: A reward is a single feedback signal observed at one timestep whereas the return is a discounted sum of rewards. Presumably, the next sentence should be talking about rewards, not returns?}
%Each square in the grid may contain a diamond or a fire; transitioning into a diamond square \emph{adds} 20 to the agent's return, while running into a fire square \emph{subtracts} 100 from the agent's return. Moreover, each step the agent takes adds -1 to the agent's return.
Each square in the grid may contain a diamond or a fire; transitioning into a diamond square gives the agent a reward of +20, while running into a fire square gives the agent a reward of -100. Moreover, each step the agent takes incurs a reward of -1.
An episode ends either when the agent transitions into a specific goal square, or when the episode length passes a predetermined constant threshold. We create two such grid world settings: GridWorld-v1, a smaller grid with a maximum episode length of 50, and GridWorld-v2, a larger grid with a maximum episode length of 100. Intuitively, an optimal agent will accumulate diamonds as quickly as possible while avoiding fire, and eventually navigate to the goal square to finish the episode. 

The credit-assignment challenge in this domain arises from the fact that all rewards are delayed until the very end of an episode; in other words, the sum of the rewards received for each individual transition is given as a single cumulative reward at the end of the episode, and all intermediate rewards given to the agent are zero.
%\dnote{TODO: this could be a very minor thing, but what is the agent's state representation here? Does this temporal delay technically break the Markov property of the reward function?}
We note that delaying the reward in this manner violates the Markov property of the MDP reward function and, technically, would be better modeled as a partially-observable MDP (POMDP)~\citep{kaelbling1998planning} where rewards are a function of a latent underlying state not observable to the agent. However, the optimal policy for this particular POMDP (and future delayed-reward settings we will introduce) can still be learned purely as a function of the partial observation, obviating the need for considering the underlying latent state or redefining states around the full trajectory. We also note that this approach of delaying rewards in an MDP while keeping the rest of the structure the same has been studied extensively in prior work~\citep{arjona2019rudder,hung2019optimizing, ren2022learning, gangwani2020learning}.

% of the reward function since it is now a function of the full trajectory and not the state, we note that the optimal policy for this environment can be learned purely as a function of the current state and not the full trajectory.

This delayed reward setting forces a good agent to disentangle the fact that collecting diamonds results in positive rewards while moving through fire results in bad rewards by assigning credit to individual actions within an episode. As we demonstrate in the following sections, this is a challenging task for algorithms which do not perform explicit credit assignment.  \vspace{1mm}

\textbf{LunarLander:} To evaluate methods on a more complex discrete action environment, we turn to the LunarLander-v2 benchmark from OpenAI Gym \cite{brockman2016openai}, in which an agent attempts to successfully land a spaceship within a specified zone. Again, to impose a credit-assignment challenge in this setting, the environment is modified such that rewards are delayed to the end of the episode. We additionally evaluate the agent with maximum episode lengths of 500 and the default length of 1000. The environment is considered ``solved'' when the agent achieves a return of 200.\vspace{1mm}

\textbf{MuJoCo Gym Suite:} Lastly, we benchmark methods on complex continuous control settings using four environments from the MuJoCo suite~\cite{6386109}: HalfCheetah, Humanoid, Walker2d, and Swimmer. Again, all rewards are delayed to episode termination and, additionally, episodes in all environments are truncated in length to a maximum of 100 steps; we additionally evaluate all methods in HalfCheetah with a maximum episode length of 50, as part of an ablation study.

\begin{figure*}[t!]
\centering
\begin{subfigure}{0.33\textwidth}
  \centering
  \includegraphics[width=1.0\linewidth]{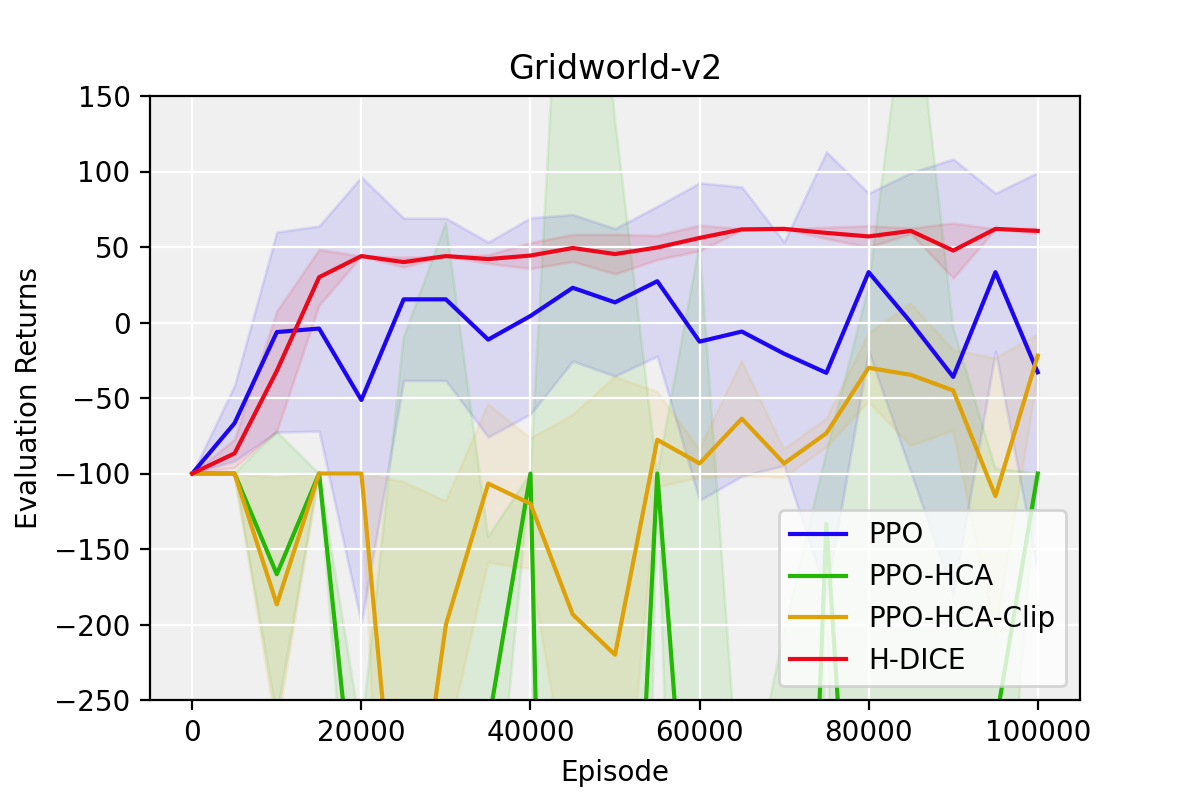}
  % \caption{A subfigure}
  \label{fig:sub2_1}
\end{subfigure}
\begin{subfigure}{0.33\textwidth}
  \centering
  \includegraphics[width=1.0\linewidth]{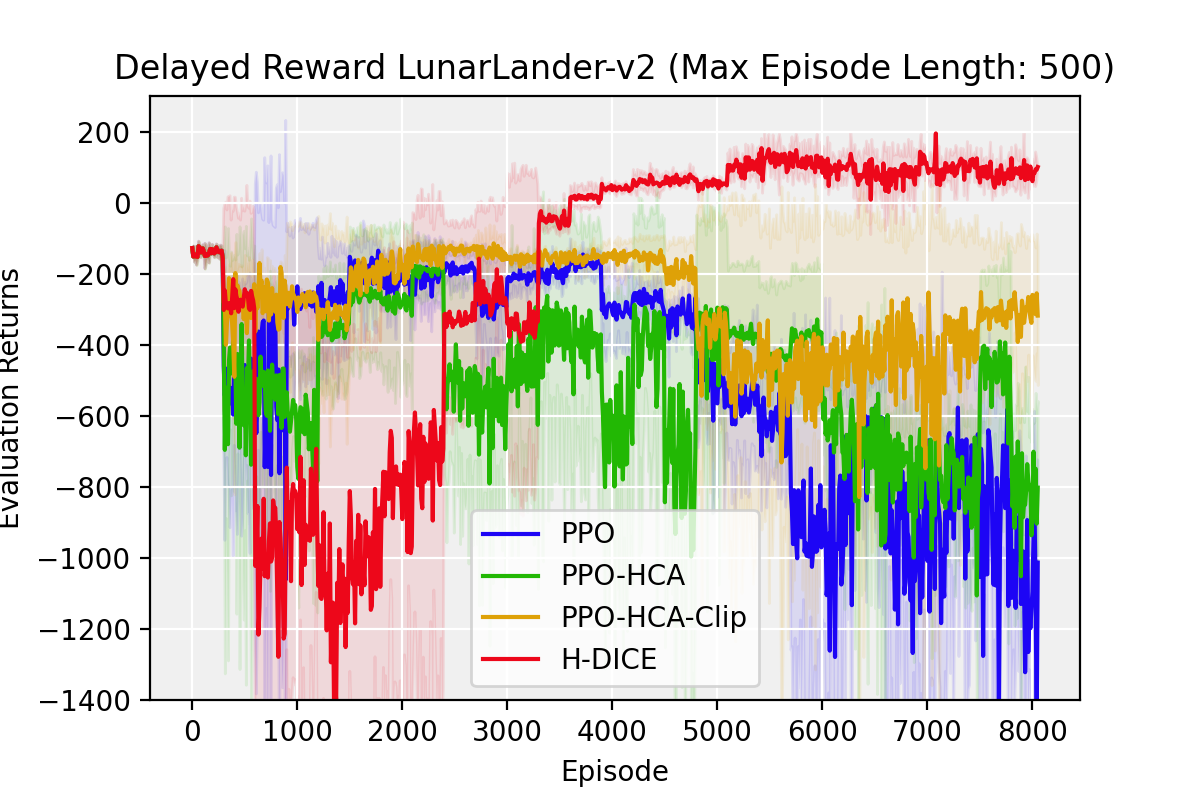}
  % \caption{A subfigure}
  \label{fig:sub2_2}
\end{subfigure}%
\begin{subfigure}{0.33\textwidth}
  \centering
  \includegraphics[width=1.0\linewidth]{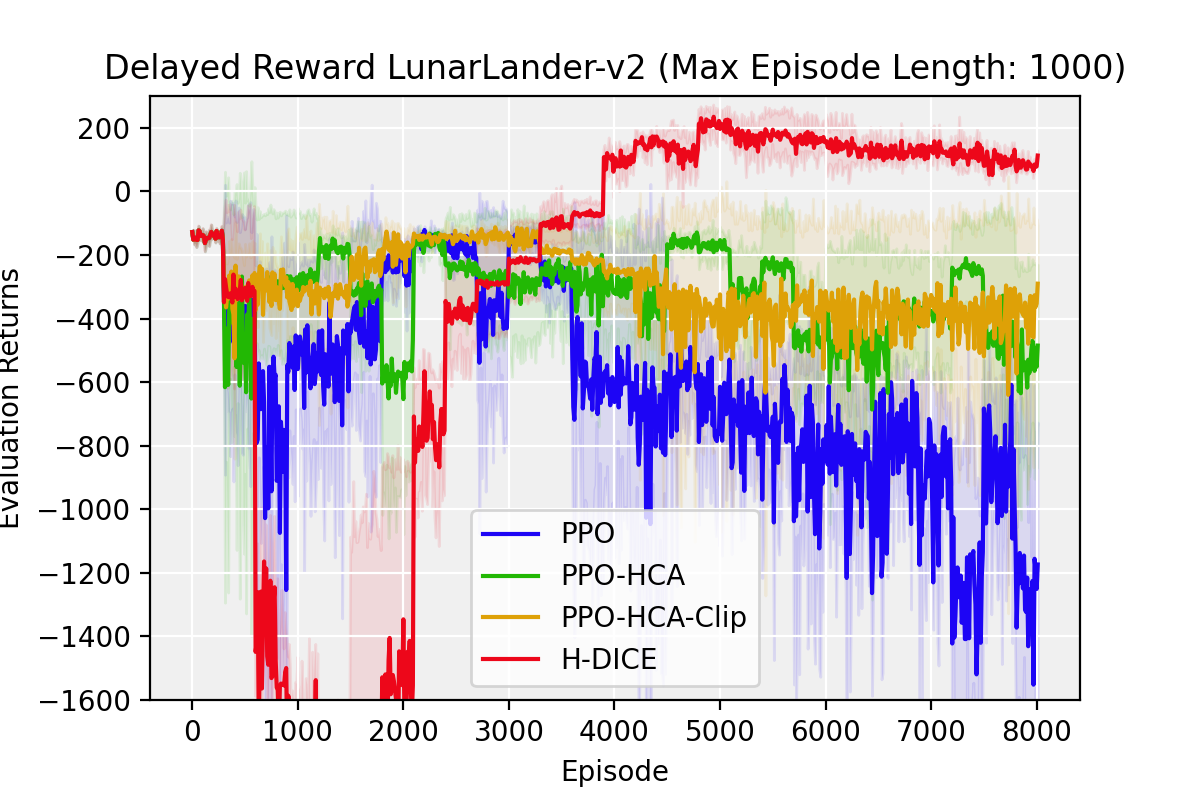}
  % \caption{A subfigure}
  \label{fig:sub2_3}
\end{subfigure}
% \vspace{-20pt}
\caption{Training curves of H-DICE and baseline methods in Gridworld-v2 (left), and delayed reward LunarLander-v2 with maximum episode lengths of 500 (middle) and 1000 (right). The LunarLander-v2 environment is considered ``solved'' when returns surpass 200. All three environments utilize a discrete action space. H-DICE outperforms baseline methods in all three environments.}
\label{fig:LunarLander}
\end{figure*}

\subsection{Baselines}
We compare \methodname{} to three baseline approaches:

% \begin{itemize}
 \textbf{Vanilla PPO:} Our first baseline is PPO with GAE-advantage calculation \cite{schulman2017proximal, schulman2018highdimensional}, a method which does not perform explicit credit-assignment. \vspace{1mm}
    
\textbf{PPO-HCA:} This baseline calculates the advantage as proposed by return-conditioned HCA \cite{NEURIPS2019_195f1538} instead of the typical Generalized Advantage Estimate \cite{schulman2018highdimensional} used in PPO, but computes the hindsight ratio directly by simply dividing the policy and hindsight policy likelihoods. As shown in the results, this method of computing hindsight ratios is highly unstable and results in poor policy performance. \vspace{1mm} 
    
\textbf{PPO-HCA-Clip:} This baseline is identical to PPO-HCA, but attempts to stabilize the hindsight ratio (which is still computed by dividing the policy and hindsight policy ratios) by clipping the ratio to be between zero and one. We choose these bounds since we find smaller clip ranges improves performance and since the density ratios in \methodname{} are between zero and one by virtue of the fact that, by default when $C=1$, the hindsight DICE model and return predictor models output values between zero and one.
% \end{itemize}

\subsection{Experimental Setup}
We tune \methodname{} and each baseline method via a limited grid search over critical hyperparameters. 
Specifically, for each domain, we search over the entropy bonus coefficient and the amount of data collected before each policy update for each method.
For HCA methods and \methodname{}, we additionally tune the number of epochs for updating the hindsight model, return predictor, and DICE model.
Lastly, for the vanilla PPO baseline, we also tune the coefficient on the value-function loss. 
We refer the reader to Appendix \ref{app:training} for a detailed account of hyperparameters used.
All results are averaged over three seeds; the solid lines in the plots correspond to the mean return and the shaded region to one standard deviation from the mean. 

% We report plots in which the shaded region correspond to the minimum and maximum returns over seeds in Appendix Section \ref{App: MinMax_Curves} and these plots clearly indicate that \methodname{} outperforms baselines consistently across seeds.

% \begin{figure}[h!]
%     \vspace{-10pt}
%     \centering
%     \includegraphics[width=0.35\textwidth]{images/updated_plots/Gridworld-v2_std_v6.png} \hspace{2pt}
%     \vspace{-9pt} 
%     \caption{Training curves of H-DICE and baseline methods on Gridworld-v2. H-DICE achieves higher final returns than baseline methods across all seeds, while baseline methods strugle with unstable and inefficient learning.} \vspace{-10pt}
%     \label{fig:Gridworld-v2}
% \end{figure}

\vspace{-2pt}
\subsection{How does \methodname{} perform compared to baselines on complex continuous control and discrete control tasks in the delayed reward setting}
\label{sec:performance}
\begin{figure*}[t!]
\centering

\begin{subfigure}{0.33\textwidth}
  \centering
  \includegraphics[width=1.0\linewidth]{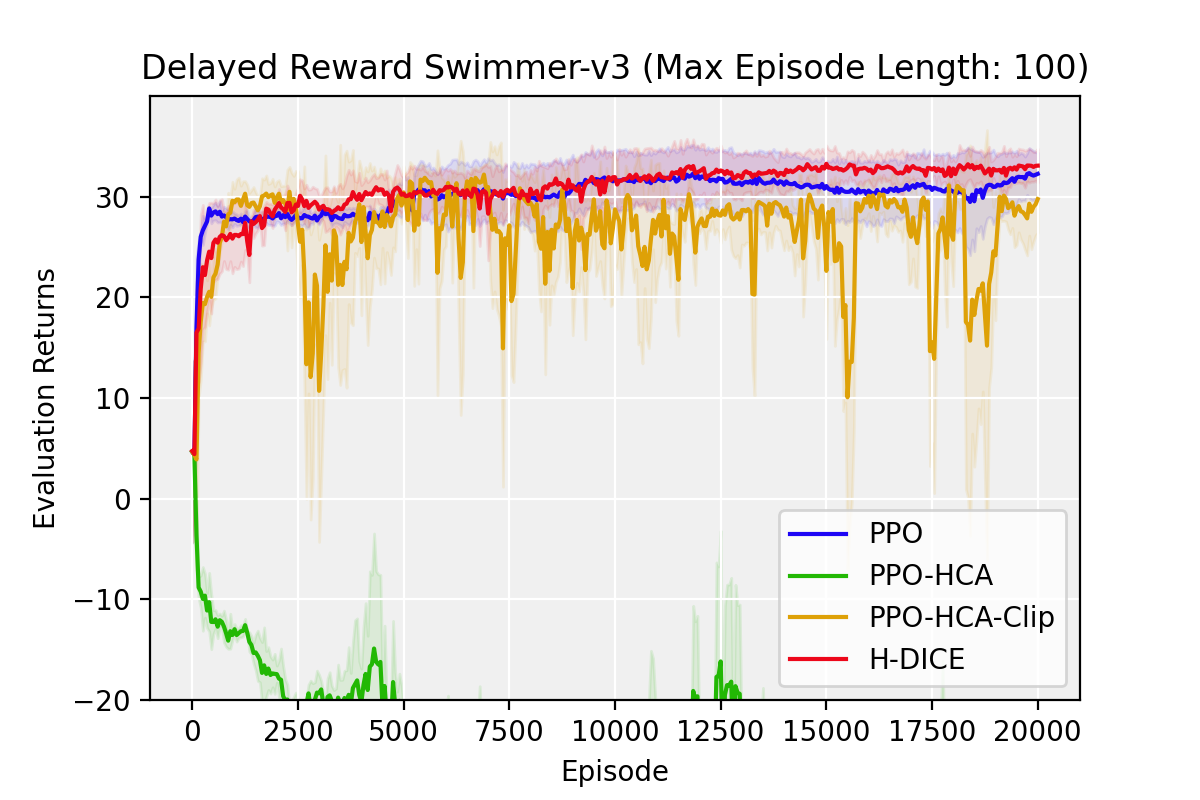}
  % \caption{A subfigure}
  \label{fig:sub3_3}
\end{subfigure}%
\begin{subfigure}{0.33\textwidth}
  \centering
  \includegraphics[width=1.0\linewidth]{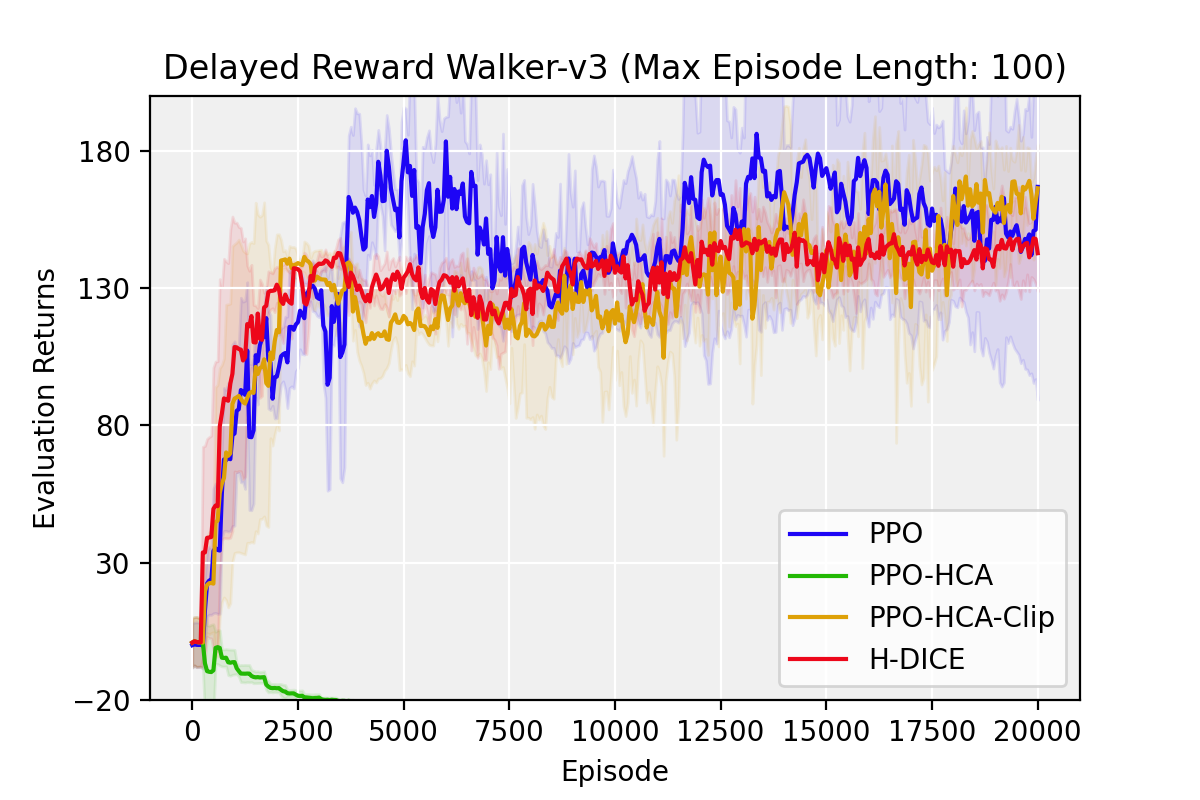}
  % \caption{A subfigure}
  \label{fig:sub3_4}
\end{subfigure}
\begin{subfigure}{0.33\textwidth}
  \centering
  \includegraphics[width=1.0\linewidth]{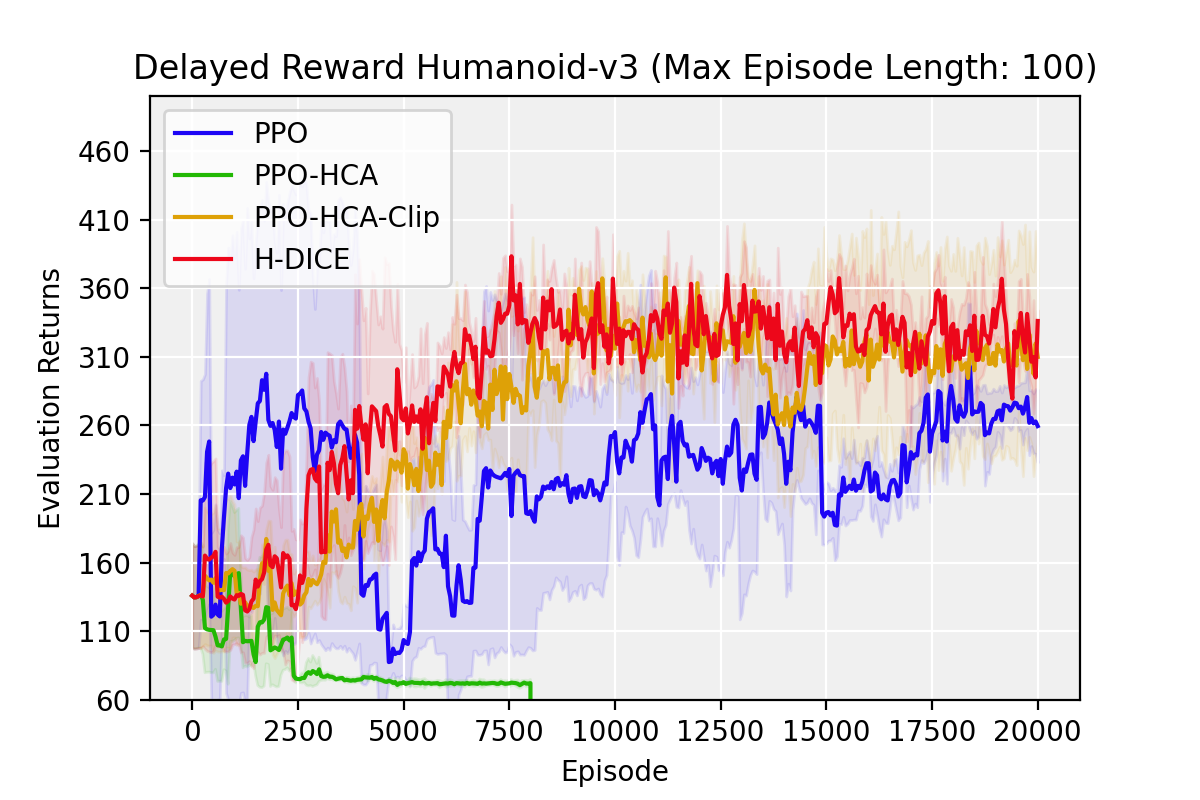}
  % \caption{A subfigure}
  \label{fig:sub3_5}
\end{subfigure}
% \vspace{-15pt}
\caption{Training curves of H-DICE and baseline methods in Half-Cheetah with a maximum episode length of 50 and 100, and in other MuJoCo environments with a maximum episode length of 100. H-DICE outperforms baseline methods in the more difficult Half-Cheetah and Humanoid environments, and performs similarly to PPO and PPO-HCA-Clip on Swimmer and Walker.}
\vspace{-10pt}
\label{fig:Mujoco}
\end{figure*}

\vspace{-2pt}
Figures \ref{fig:intro}, \ref{fig:LunarLander} and \ref{fig:Mujoco} display learning curves of \methodname{} and baseline algorithms in a variety of different discrete and continuous control environments. The vertical axis denotes the returns achieved by the agent averaged over 10 evaluation episodes whereas the horizontal axis denotes the number of episodes elapsed since the beginning of training. 
%\dnote{TODO: if we were doing value-based RL, I would interpret an evaluation episode as an episode where we use the greedy policy of our $Q$-function approximation. I'm not sure what this means in the context of a policy search method.}

\methodname{} achieves higher final returns and converges faster than baseline methods in the vast majority of scenarios.
PPO notably struggles to achieve high returns in many of these domains and is typically outperformed by \methodname{} in final returns and learning efficiency. The exception to this is in the Swimmer and Walker environments, where PPO, PPO-HCA-Clip, and \methodname{} perform similarly.
This demonstrates the necessity for explicit credit-assignment in settings where the lack of immediate reward feedback requires reasoning about actions were responsible for generating the final return.
As illustrated by the poor results of PPO-HCA across domains, computing hindsight ratios directly by dividing the policy and hindsight policy likelihoods results in instability issues and leads to suboptimal performance. The variance across seeds is extremely high, and the achieved returns oscillate drastically particularly in the GridWorld and LunarLander settings. Stabilizing the hindsight ratios via clipping, the strategy used by PPO-HCA-Clip, alleviates this instability in most settings, as evidenced by both the improvement in returns and the reduction in variance as compared to PPO-HCA. As it still explicitly models credit-assignment, PPO-HCA-Clip also outperforms PPO in several settings, but is clearly outperformed by \methodname{} across most environments, demonstrating the need for more sophisticated methods for stable computation of the hindsight ratio.

We also note several interesting trends in the relationship between episode length and method performance. As illustrated by the final achieved rewards in the LunarLander and Half-Cheetah plots in Figures \ref{fig:LunarLander} and \ref{fig:Mujoco} respectively, the difference in performance between \methodname{} and baseline methods (particularly PPO) is more pronounced when the maximum episode length is longer. 
This is likely explained by the fact that credit assignment becomes increasingly important over longer horizons, as the effects of earlier actions can manifest much later in the episode, and because the temporal distance between when actions are executed and when the agent receives a nonzero reward is greater.

\begin{figure*}[t!]
\setcounter{figure}{1}
\centering
\begin{subfigure}{0.35\textwidth}
  \centering
  \includegraphics[width=1.0\linewidth]{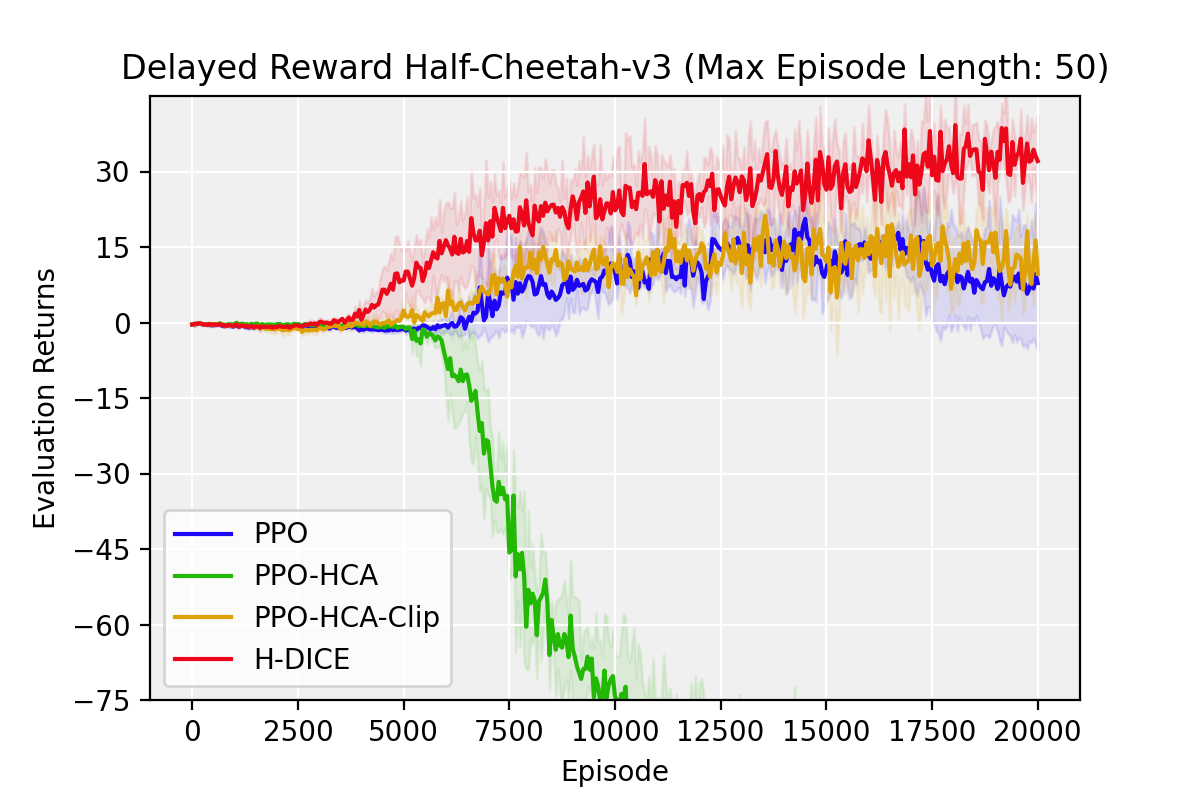}
  % \caption{A subfigure}
  \label{fig:sub3_1}
\end{subfigure}%
\begin{subfigure}{0.35\textwidth}
  \centering
  \includegraphics[width=1.0\linewidth]{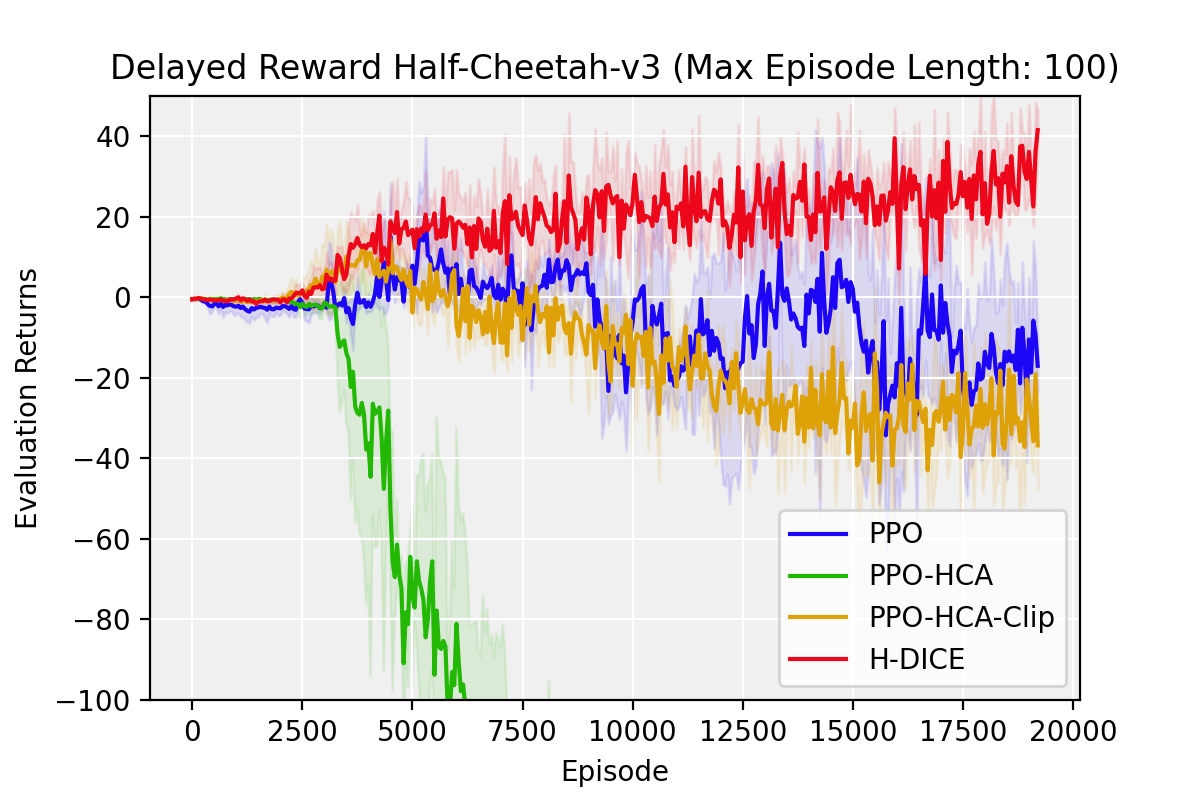}
  % \caption{A subfigure}
  \label{fig:sub3_2}
\end{subfigure}
\vspace{-10pt}
\label{fig:HalfCheetah}
\end{figure*}

\subsection{Do the hindsight ratios learned by \methodname{} encode meaningful credit assignment trends?}
\label{sec:interpretability}
\vspace{-2pt}
To answer this question, we take a fully trained \methodname{} model on GridWorld-v1 and analyze the learned $\pi_\theta$ and $h_{\pi_\theta}$ values at a critical state in a trajectory -- namely, when the agent is in the position  shown in Figure \ref{fig:intro}.
In this position, there are two interesting actions the agent can take: left or right, each of which leads to vastly different outcomes. Conditioned on the achieved final return, the agent must assign credit back to the action taken at this step to determine if it was culpable.

%\dnote{TODO: I've started noticing more conflations between ``rewards'' and ``returns'' below. Important to be clear about this; returns are ultimately what get used to condition the hindsight policy, not rewards.}
In the table below, we show values corresponding to the (hypothetical) action taken by the agent at this state ($s$), the (hypothetical) achieved final return ($z$), the probability of taking that action under policy $\pi_\theta$ at this state and the hindsight probability, $h^{\pi_\theta}(a \mid s, z)$, of choosing this action at this state given the final return and the current policy.
From these values, we also show the naive ratio calculation of $\frac{\pi_\theta(a \mid s)}{h^{\pi_\theta}_\omega(a \mid s, z)}$ and the same ratio obtained via \methodname{} using Equation \ref{eq:H-dice-final}.

\begin{table}[H]
\caption{Table showing the learned values from \methodname{} at the end of training in GridWorld-v1 and the corresponding ratios obtained from naive division and \methodname{}.}
\centering
\begin{tabular}{|c|c|c|c|c|c|}
\hline
\textbf{Action (a)} & \textbf{Return (z)} & $\pi_\theta(a \mid s)$ & $h^{\pi_\theta}_\omega(a \mid s)$ & \textbf{Direct Ratio} & \textbf{\methodname{} Ratio} \\ \hline \hline
\textsc{Left}                & -100                & 0.002                           & 0.349                                &  0.006                & 0.139                                     \\ \hline
\textsc{Left}                & 69                  & 0.002                           & 0.163                                                                 & 0.013                & 0.015                                    \\ \hline
\textsc{Right}               & -100                & 0.845                           & 0.210                                                                 & 4.019                & 0.139                                     \\ \hline
\textsc{Right}               & 69                  & 0.845                           & 0.434                                                                 & 1.949                & 0.014                                   \\ \hline
\end{tabular}
\label{tab:interpretability}
\end{table}

\vspace{-2pt}

Going \textsc{Left} from the given state would transition the agent into the fire and likely lead to a return of approximately -100 at the end of the episode. 
In contrast, moving \textsc{Right} would enable the agent obtain a diamond and is likely lead to a high positive return, such as 69, at the end of the episode.
The probability of selecting these respective actions and the corresponding hindsight probabilities are accurately captured in columns 3 and 4 of Table \ref{tab:interpretability}.
Specifically, note that the probability of choosing the action \textsc{Left} when we achieve a reward of -100 is a lot higher than if the reward were 69. A similar trend holds true for the \textsc{Right} action.
In short, we observe that the hindsight model qualitatively models the conditional distribution of actions given returns accurately. 

% \anote{is paragraph this needed? I don't think it conveys that much important information.} While the output of the density model is harder to interpret, the output of the return model is as expected as it indicates that the probability of achieving a reward of 69 is much more likely than the probability of achieving a reward of -100 from that state. 
% This is expected for a fully trained model.

The final two columns of the table show the direct $\frac{\pi_\theta(a \mid s)}{h^{\pi_\theta}_\omega(a \mid s, z)}$ ratio values which are computed by simply dividing the two required probabilities, and the approximated ratio obtained via \methodname{} using Equation \ref{eq:H-dice-final}.
We observe that the direct ratio -- which is a more accurate estimate of the hindsight ratio of interest -- has much larger variation: values fluctuate from 0.006 to 4.019 even in a simple environment such as GridWorld-v1. 
On the other hand, ratios computed using \methodname{} are much lower variation and only lie between 0.014 and 0.139 while still mirroring the trends of the direct ratio. For instance, for the \textsc{Right} action, the direct ratio is 4.019 and 1.949 for returns of -100 and 69 respectively. The corresponding \methodname{} ratios are 0.139 and 0.014, which are correlated to the true ratios. We posit that this ability to capture the direct ratio's trends while reducing variance is a core reason for the improved performance of \methodname{} compared to PPO-HCA. Note, however, that there are times when inaccuracies in auxiliary models prevent \methodname{} from reflecting the trends of the direct ratio; an instance of such discrepancies can be seen in the ratios computed for the \textsc{Left} action. Nevertheless, as backed by the results in Section \ref{sec:exps}, the stability which results from \methodname{} is crucial to achieving strong empirical results in challenging credit-assignment settings. Lastly, while simple settings such as GridWorld-v1 enable us to interpret model outputs at the level of individual state-action pairs, we suspect that these issues are only exacerbated in more complex environments, further underscoring the importance of credit assignment in deep RL.

\subsection{Can we update the auxiliary models with off-policy schedules?}
\label{sec:off_policy}
\vspace{-2pt}

A potential drawback of \methodname{} stems from the fact that four models must be updated at once to learn the hindsight ratio, which can lead to greater training overhead and challenges with hyperparameter tuning. 
This issue is exacerbated by the fact that all these models must be updated in an on-policy fashion as per the methodology detailed in Section \ref{sec: HDICE}. This overhead could be alleviated if the auxiliary models could be updated with off-policy schedules, i.e., less frequently than the policy network, thereby allowing them to be trained with data collected from previous versions of the policy.
To this end, we aim to study the impact of updating the auxiliary models with various off-policy schedules on final performance. 

\begin{figure*}[t!]
\centering
\begin{subfigure}{0.45\textwidth}
  \centering
  \includegraphics[width=1.0\linewidth]{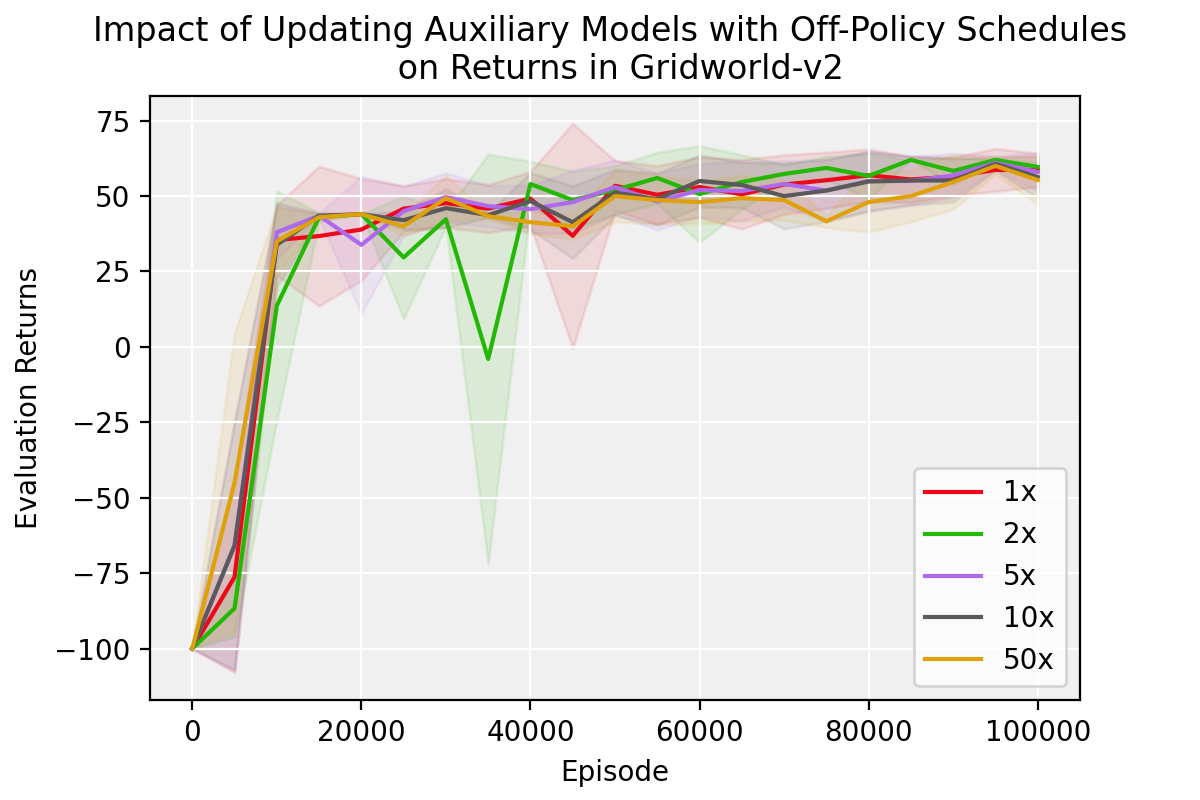}
  % \caption{A subfigure}
  \label{fig:sub4_1}
\end{subfigure}
\begin{subfigure}{0.45\textwidth}
  \centering
  \includegraphics[width=1.0\linewidth]{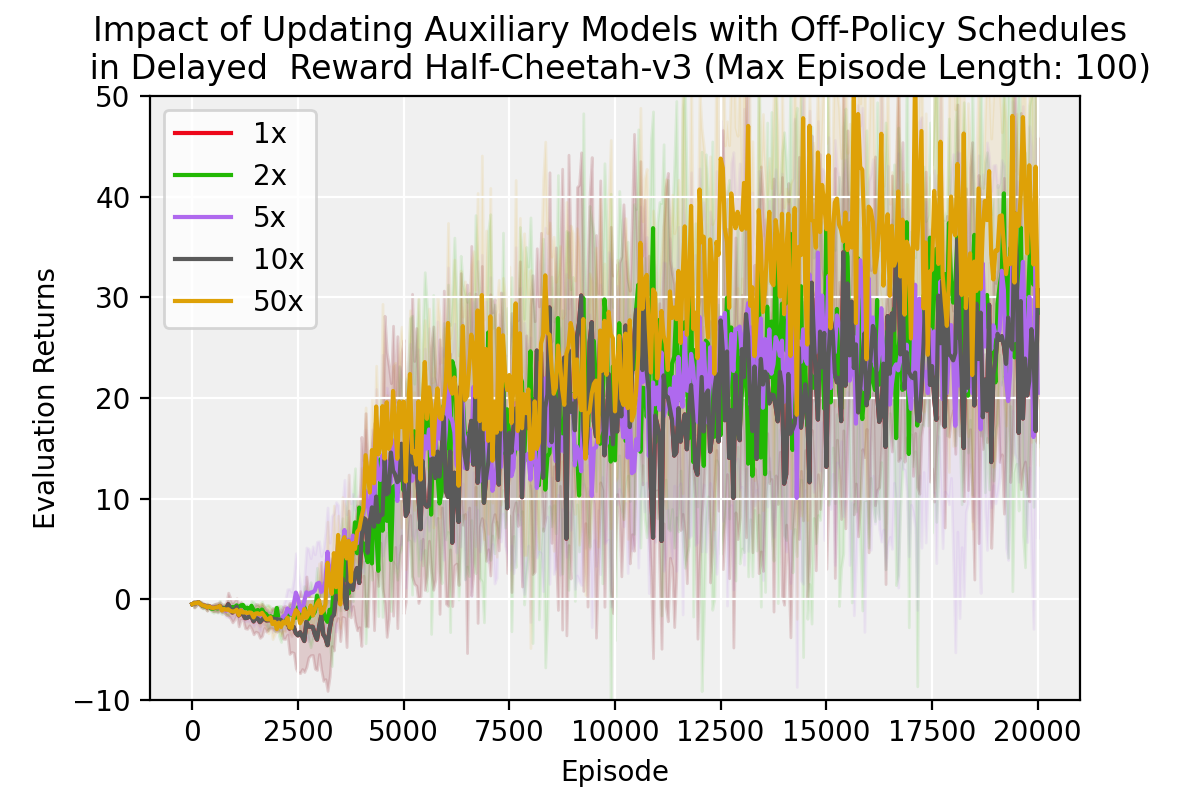}
  % \caption{A subfigure}
  \label{fig:sub4_2}
\end{subfigure}%
\caption{Training curves of varying off policy schedules on GridWorld-v2 and HalfCheetah. There is almost no performance loss when updating with off-policy schedules and in some cases we can even see improvements.}
\label{fig:op}
\end{figure*}

We perform an ablation on GridWorld-v2 and HalfCheetah environment where we progressively slow down the updates to the auxiliary models relative to the update frequency of the policy.
In the plots shown in Figure \ref{fig:op}, the curves labeled as $N$x indicate that the auxiliary models are all updated $N$ times slower than the policy model, and as a result with $N$ times more data (which will contain trajectories collected from previous iterations of the policy). Note that auxiliary models' weights are still reset before their updates.

As seen in Figure \ref{fig:op}, there is hardly any performance loss when updating with off-policy schedules. In some instances, for example when the auxiliary models are updated $50$x less frequently than the policy in HalfCheetah, there see a small improvement over updating all models in lockstep.
We conjecture that this similarity in performance may be a result of the trade-off between updating models in a fully on-policy fashion but limiting the volume of training data for the auxiliary models, and being more off-policy which enables models to be trained with more data.

% being perfectly on-policy and giving the auxiliary models more data to train, thereby giving them better generalizing capabilities.

Finally, we note that updating the auxiliary models in an off-policy fashion strays away from the theoretical justification for \methodname{} developed in Section \ref{sec: HDICE}, but enables \methodname{} to be computationally efficient with little to no cost to empirical performance. 

% is not supported by our math but is a purely empirical study that may help improve computational efficiency at little to no cost to performance.

\section{Related Work}
\label{sec:related}

The classic workhorse for addressing credit assignment in reinforcement learning has been the eligibility trace~\citep{klopf1972brain,sutton1984temporal,sutton1988learning} which awards credit in a geometrically-decaying fashion to states or state-action pairs on the basis of temporal recency to surprising events, as measured by the temporal-difference (TD) error. While these traditional approaches laid the groundwork for the field, the scale of modern problems makes them woefully insufficient and impractical. More recently, a new class of emphatic TD methods have emerged~\citep{sutton2016emphatic,hallak2016generalized,yu2015convergence,anand2021preferential,chelu2022selective} which includes a so-called interest function that enables more nuanced and non-linear attribution of credit for incident states along a sampled trajectory. The theory surrounding these methods, however, operate in general terms without concrete specification of such an interest function. Since scalable methods for leveraging interest functions when learning option initiation sets exist~\citep{sutton1999between}, future work may find it fruitful to learn interest functions using our \methodname{} approach in the context of emphatic TD methods. Finally, recent work on expected eligibility traces~\citep{van2021expected,chelu2022selective} acknowledges the trajectory as a random variable and aims to learn eligibility traces that award credit to those realizations of the trajectory that were not observed but could have occurred under the given behavior policy; indeed, the hindsight policy studied in this work can be shown to emerge through an information-theoretic treatment of credit assignment where such randomness is accounted for~\citep{arumugam2021information}. Still, expected eligibility traces suffer from the same temporal recency heuristic that plagues traditional eligibility traces, whereas the hindsight policy enables non-linear re-weighting of on-policy data.

Our work extends the hindsight credit assignment (HCA) work of \citet{NEURIPS2019_195f1538}, rectifying instabilities in using hindsight policies for credit assignment. While \citet{alipov2021towards} find that an alternative form of clipping applied to hindsight probabilities can improve stability in Atari games, our experiments demonstrate across a broad spectrum of domains that clipping can greatly stall the speed at which learning progresses with HCA. In contrast, an alternative and scalable approach to tackling credit assignment emerges through the concept of return redistribution~\citep{arjona2019rudder,hung2019optimizing, ren2022learning, gangwani2020learning}. Unfortunately, practical agents employing return redistribution rely on recurrent network architectures whose performance and susceptibility to vanishing gradients rises as the problem horizon increases. 
Although methods such as \citet{gangwani2020learning} and \citet{ren2022learning} do not use recurrent network architectures, they use simplistic methods such as uniform reward decomposition or least squares reward decomposition, which do not necessarily have to assign any meaningful or interpretable credit to the transitions. 
In contrast, hindsight policies for data re-weighting retain an elegant idea from classic temporal-difference learning; namely, that local information can be stitched together across timesteps to yield globally effective solutions.
There have also been recent approaches that explore the relationship between causality and counterfactuals to the RL credit assignment problem~\citep{mesnard2021counterfactual, buesing2018woulda}; while applying such powerful tools from causal inference does seem like a plausible and even promising avenue for credit assignment in RL, the main drawback to these existing approaches is assuming a perfect causal model of the environment and, with greater limitation, the ability to perform perfect counterfactual inference queries where one factor of variation is manipulated \textit{ceteris paribus} or while keeping everything else fixed. In contrast, \methodname{} is theoretically grounded within the framework of importance sampling~\citep{NEURIPS2019_195f1538} without such cumbersome assumptions.

\section{Conclusion}
\label{sec:conc}
In this work, we presented \methodname{}, a novel algorithm for \emph{stable} credit assignment in deep reinforcement learning. Building on the Hindsight Credit Assignment (HCA) framework introduced by \citet{NEURIPS2019_195f1538} and noting the instabilities that arise when directly estimating hindsight density ratios, we take inspiration from methods in the off-policy evaluation literature \cite{nachum2019dualdice} to develop a method which computes the quantities needed to perform credit assignment via HCA in a stable fashion. We showed that in challenging benchmarks with delayed rewards, algorithms that do not assign credit or do so in a naive fashion struggle to perform well while \methodname{} converges to higher returns much faster and more consistently than baseline methods.

Our work utilizes a formulation of HCA in which the hindsight policy is conditioned on future outcome information in the form of returns. While this performs well in delayed-reward settings, it requires a diversity of returns to be observed during training in order for the hindsight policy to model something meaningful beyond the policy itself. Hence, it is likely that \methodname{} will struggle in sparse-reward settings where returns are either zero or one due to the lack of diversity in observed returns. To address this orthogonal exploration challenge in future work, we hope to extend this work to state-conditioned HCA, a formulation in which the hindsight policy is conditioned on future states in a trajectory rather than the return. Anticipating that it is easier to achieve diversity in state visitation rather than a wide range of potential returns, we suspect this variant is likely to fare better in sparse-reward, language conditioned and goal-reaching settings.

While we conducted some preliminary experiments in an effort to extend \methodname{} to operate in an off-policy fashion, the core algorithm is still an on-policy algorithm built atop conventional policy-gradient methods.
An interesting line of future-work will be to extend credit assignment to off-policy algorithms such as DQN \cite{mnih2013playing} and SAC \cite{haarnoja2018soft} which are typically more data efficient than on-policy algorithms. By bridging ideas from OPE to address the credit assignment challenge, we hope to open the door for further symbiotic progress moving forward on both fronts. Developing credit assignment methods that are able to perform expressive, yet stable updates will be crucial for efficient deep RL, in a variety of settings in which meaningful reward feedback is rarely received. 

\bibliography{references}
\bibliographystyle{plainnat}

%%%%%%%%%%%%%%%%%%%%%%%%%%%%%%%%%%%%%%%%%%%%%%%%%%%%%%%%%%%%%%%%%%%%%%%%%%%%%%%
%%%%%%%%%%%%%%%%%%%%%%%%%%%%%%%%%%%%%%%%%%%%%%%%%%%%%%%%%%%%%%%%%%%%%%%%%%%%%%%
% APPENDIX
%%%%%%%%%%%%%%%%%%%%%%%%%%%%%%%%%%%%%%%%%%%%%%%%%%%%%%%%%%%%%%%%%%%%%%%%%%%%%%%
%%%%%%%%%%%%%%%%%%%%%%%%%%%%%%%%%%%%%%%%%%%%%%%%%%%%%%%%%%%%%%%%%%%%%%%%%%%%%%%
%\newpage
\appendix

\onecolumn

\section{Solution to the optimization in Equation \ref{eq:h-dice-opt}}
\label{app:math}
In this section, we show that the solution to the following optimization introduced in Section \ref{sec: HDICE}:

$$\min_{\phi: S \times A \times Z \rightarrow \mathcal{C}} \frac{1}{2}\expect_{(s, a, z) \sim \mc{D}^{{h}^{\pi_\theta}}}[\phi(s, a, z)^2] - \expect_{\substack{(s, a) \sim d^{\pi_\theta} \\ z \sim \psi(z)}}\left[ \phi(s, a, z) \right]$$

is given by 

$$\phi^{*}(s, a, z) = \frac{d^{\pi_\theta}(s, a) \psi(z)}{\mc{D}^{h^{\pi_\theta}}(s, a, z)}$$

One can also consult the salient exposition of \citet{nachum2019dualdice} for an alternative rationale and calculation but specialized to their OPE setting instead of our credit assignment focus. We begin by expanding the expectations inside the objective:

\begin{align*}
    &\frac{1}{2}\expect_{(s, a, z) \sim \mc{D}^{{h}^{\pi_\theta}}}[\phi(s, a, z)^2] - \expect_{\substack{(s, a) \sim d^{\pi_\theta} \\ z \sim \psi(z)}}\left[ \phi(s, a, z)\right] \\ 
    &= \frac{1}{2}\hspace{-7mm}\int\displaylimits_{s \in S, a \in A, z \in Z} \hspace{-7mm}\mc{D}^{h^{\pi_\theta}}(s, a, z)\phi(s, a, z)^2 ds da dz - \hspace{-7mm}\int\displaylimits_{s \in S, a \in A, z \in Z}\hspace{-7mm}d^{\pi_\theta}(s, a)\psi(z)\phi(s, a, z)ds da dz \\ 
    &= \hspace{-7mm}\int\displaylimits_{s \in S, a \in A, z \in Z}\hspace{-7mm}\left(\frac{1}{2}\mc{D}^{h^{\pi_\theta}}(s, a, z)\phi(s, a, z)^2 - d^{\pi_\theta}(s, a)\psi(z)\phi(s, a, z)\right)dsdadz
\end{align*}

Minimizing this quantity can be achieved by enforcing first-order optimality conditions. We first differentiate the above integral using the Leibniz Integral Rule, which allows for the derivative to move inside the integral:

\begin{align*}
    &\hspace{-7mm}\int\displaylimits_{s \in S, a \in A, z \in Z}\hspace{-7mm}\frac{\partial}{\partial \phi}\left(\frac{1}{2}\mc{D}^{h^{\pi_\theta}}(s, a, z)\phi(s, a, z)^2 - d^{\pi_\theta}(s, a)\psi(z)\phi(s, a, z)\right)dsdadz \\
    &= \hspace{-7mm}\int\displaylimits_{s \in S, a \in A, z \in Z}\hspace{-7mm}\left(\mc{D}^{h^{\pi_\theta}}(s, a, z)\phi(s, a, z) - d^{\pi_\theta}(s, a)\psi(z)\right)dsdadz
\end{align*}

We now equate this expression for the derivative to zero and solve for $\phi$. 
Similar to the proof in DualDICE~\cite{nachum2019dualdice}, we look for the value of $\phi(s,a,z)$ such that this gradient is equal to zero pointwise over all state-action-return triples and get:
%This bounded integral is 0 when the integrand is 0; hence, we can equate the expression inside the parenthesis to zero, arriving at the following expression for all combinations of $s, a, z$:

$$\phi^{*}(s, a, z) = \frac{d^{\pi_\theta}(s, a) \psi(z)}{\mc{D}^{h^{\pi_\theta}}(s, a, z)}$$

which we can then write as:

\begin{align}
    &\phi^{*}(s, a, z) = \frac{d^{\pi_\theta}(s) \pi_\theta(a \mid s)\psi(z)}{\mc{D}^{h^{\pi_\theta}}(s) \chi^{\pi_\theta}(z \mid s) h^{\pi_\theta}_\omega(a \mid s,z)} \nonumber \\
    &\implies \phi^{*}(s, a, z) = \frac{\pi_\theta(a \mid s)}{\chi^{\pi_\theta}(z \mid s) h^{\pi_\theta}_\omega(a \mid s,z)}
\end{align}

The final simplification follows from the fact that $\psi(z)$ is chosen to be a uniform distribution over all possible returns; hence, $\psi(z)$ is a constant term independent of the precise realiziation $z$, therefore, can be ignored. Ignoring this constant simply changes the magnitude of all policy updates -- this is preferred over dividing by the constant, which again, introduces the risk of instability.

For instance, the policy updates are smaller in magnitude when the advantage $A(s, a) \geq 0$ because:
\begin{align*}
    &\frac{d^{\pi_\theta}(s) \pi_\theta(a \mid s)\psi(z)}{\mc{D}^{h^{\pi_\theta}}(s) \chi^{\pi_\theta}(z \mid s) h^{\pi_\theta}_\omega(a \mid s,z)} \leq \frac{d^{\pi_\theta}(s) \pi_\theta(a \mid s)}{\mc{D}^{h^{\pi_\theta}}(s) \chi^{\pi_\theta}(z \mid s) h^{\pi_\theta}_\omega(a \mid s,z)}\\
    &\implies \frac{\pi_\theta(a \mid s)\psi(z)}{ h^{\pi_\theta}_\omega(a \mid s,z)} \leq \frac{\pi_\theta(a \mid s)}{h^{\pi_\theta}_\omega(a \mid s,z)} \\
    &\implies \left(1 -\frac{\pi_\theta(a \mid s)\psi(z)}{ h^{\pi_\theta}_\omega(a \mid s,z)}\right)A(s, a) \geq \left(1 - \frac{\pi_\theta(a \mid s)}{h^{\pi_\theta}_\omega(a \mid s,z)}\right)A(s, a)
\end{align*}

and vice-versa when $A(s, a) < 0$.
% In practice, instead of sampling from a uniform distribution over all possible returns, we just pass in a dummy value, 0, as the returns to the Hindsight DICE model when calculating the second expectation. 
% This keeps things simple and we find that this works fine in our experiments.

$d^{\pi_\theta}(s) = \mc{D}^{h^{\pi_\theta}}(s)$ follows from the fact that $h^{\pi_\theta}$ is the hindsight policy learned from the state-visitation distribution of $\pi_\theta$. 
Empirically, we use the same data that we collected from the policy to train the hindsight distribution as well and we update them with the same schedule. 

This finally gives us the following expression for the hindsight ratio of interest:
\begin{align}
\label{app: final_exp}
    \frac{\pi_\theta(a \mid s)}{h^{\pi_\theta}_\omega(a \mid s,z)} = \phi^{*}(s, a, z) \chi^{\pi_\theta}(z \mid s) 
\end{align}

There are several choices for $\psi(z)$, which in principle can be any valid distribution over the returns; a good choice of $\psi(z)$ would enable easy computation of the hindsight ratio of interest. There are several potential choices for $\psi(z)$, including  $\chi^\pi(z \mid s, a)$ or $\chi^\pi(z \mid s)$.
Using the first distribution would require values of $\phi$ to be divided by values of this distribution, risking instability issues that arise when having probabilities or likelihoods as a divisor (indeed, this is the problem we aimed to avoid in the first place). Sampling returns with respect to the latter distribution is especially appealing since it conveniently cancels out with the return predictor distribution which multiples with the output of $\phi$ to obtain the hindsight ratio in Equation \ref{app: final_exp}. However, as shown in Appendix \ref{app:psi-ablation}, sampling returns from $\chi^\pi(z \mid s)$ in the second expectation empirically yields poorer performance than sampling returns from a uniform distribution. We leave other choices for $\psi(z)$ to future work.

Note that we cannot use an arbitrary distribution to model the returns in $\mc{D}^{h^{\pi_\theta}}(s,a,z)$ because we need to decompose this term into $\mc{D}^{h^{\pi_\theta}}(s)\chi^{\pi_\theta}(z \mid s)h^{\pi_\theta}(a \mid s, z)$.
Here, it is important that the returns are sampled from the policy $\pi_\theta$ and that the distribution $\chi^{\pi_\theta}$ is the distribution of returns achieved from a state \emph{under policy $\pi_\theta$}, as this is what the hindsight policy is conditioned on.

\newpage 

\section{Ablation: Impact of Hindsight DICE Model Range}
\label{app:c-ablation}

\begin{figure}[h!]
\centering
\begin{subfigure}{0.45\textwidth}
  \centering
  \includegraphics[width=0.9\linewidth]{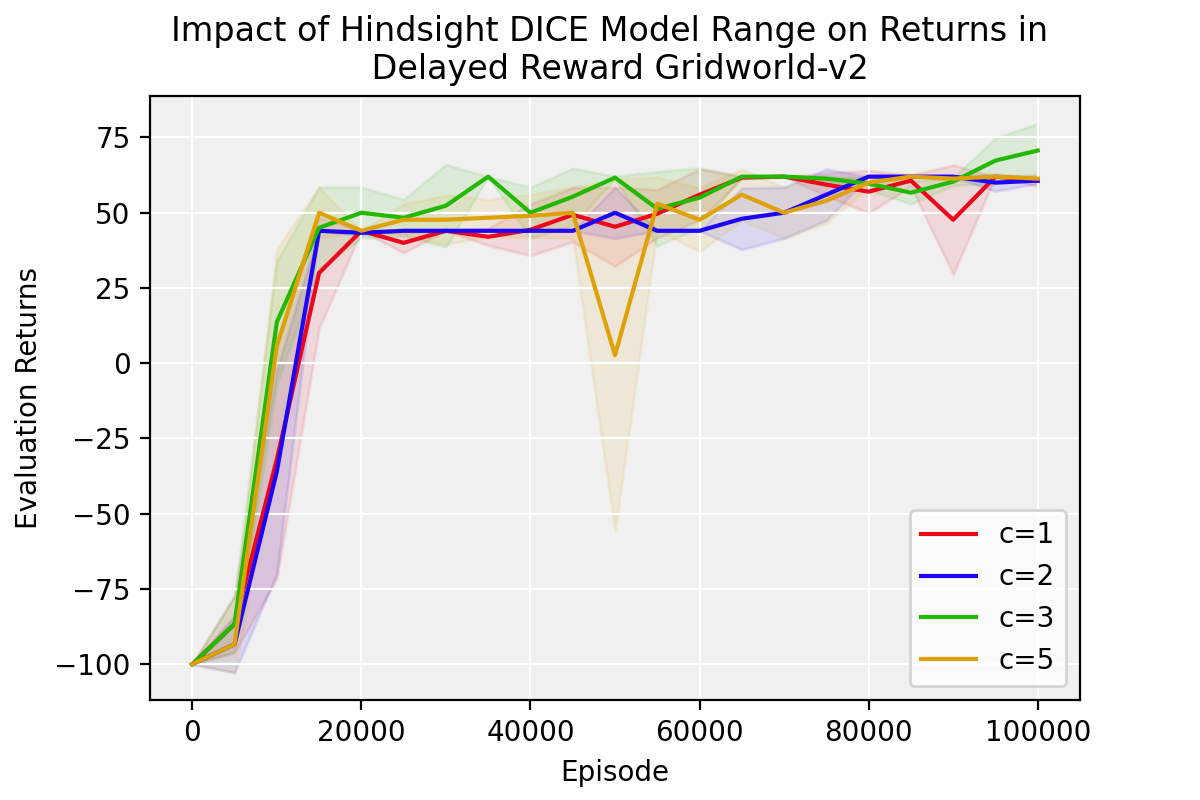}
  % \caption{A subfigure}
  \label{fig:c_ab_sub1}
\end{subfigure}%
\begin{subfigure}{0.45\textwidth}
  \centering
  \includegraphics[width=0.9\linewidth]{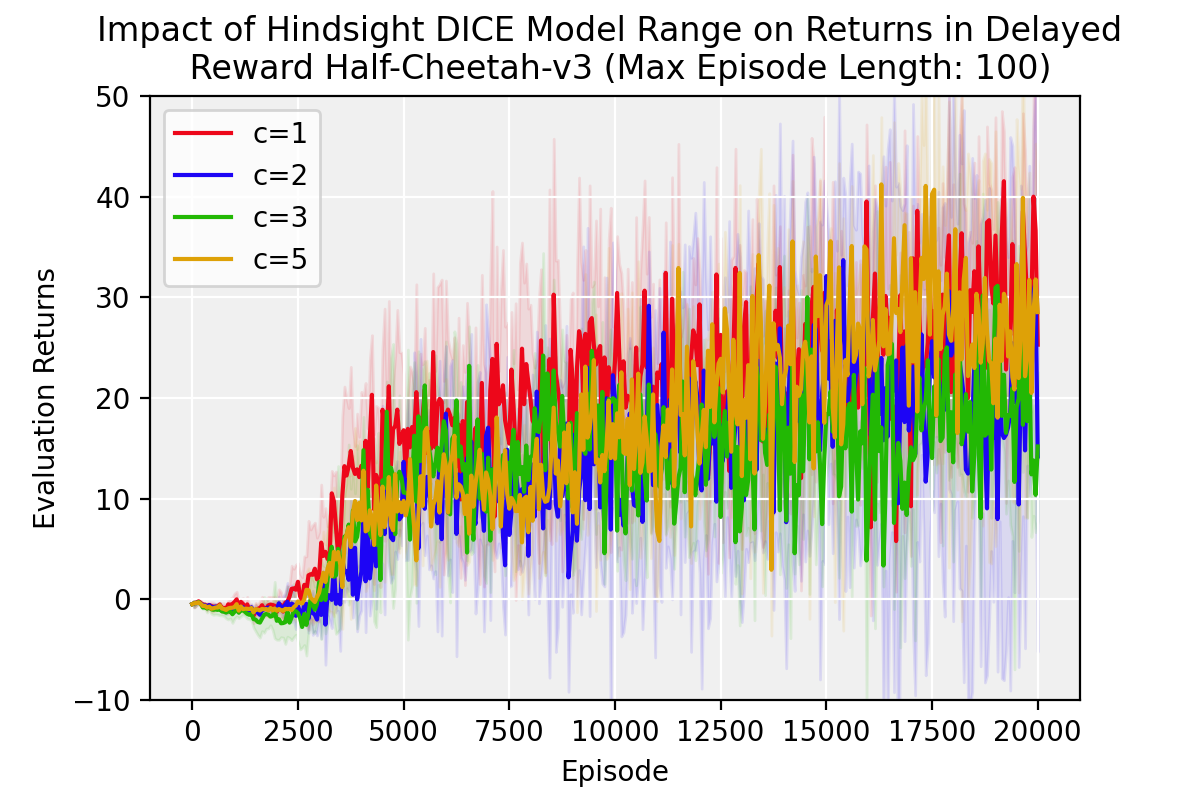}
  % \caption{A subfigure}
  \label{fig:c_ab_sub2}
\end{subfigure}
% \vspace{-15pt}
\caption{Impact of the range of the Hindsight DICE on policy performance in Gridworld-v2 (left) and Delayed Reward Half-Cheetah with a maximum episode length of 100. For a given value of $C$, the output range of the Hindsight DICE model is set to be $[0, C]$. The performance of \methodname{} is similar across choices of $C$.}
\label{fig:C-Ablation}
\end{figure}

Figure \ref{fig:C-Ablation} studies the impact of the output range of the Hindsight DICE model (which is controlled by $C$, as detailed in Section \ref{sec: HDICE}) on policy performance. All policies are trained with H-DICE, and the only axis of variation across experiments in each environment is the choice of $C$. Results are averaged across 3 seeds for each choice of $C$. Policy performance is similar across choices of $C$ in the Gridworld-v2 and delayed reward Half-Cheetah environments indicating that the choice of this hyperparameter is not very important. 

We further note that when $C=1$, the hindsight ratio calculate by \methodname{} lies in the range $[0, 1]$. 
As a result, the model is not capable of making ``double-sided'' updates, i.e., down-weighting an action at a state even when the observed returns from that state are positive (and vice versa).
This is no longer the case when $C>1$ but this boost in expressivity risks a potential cost to stability as evidenced by the results of PPO-HCA.

\section{Ablation: Choice of $\psi$ return distribution}
\label{app:psi-ablation}

\begin{figure*}[h!]
\centering

\begin{subfigure}{0.33\textwidth}
  \centering
  \includegraphics[width=0.9\linewidth]{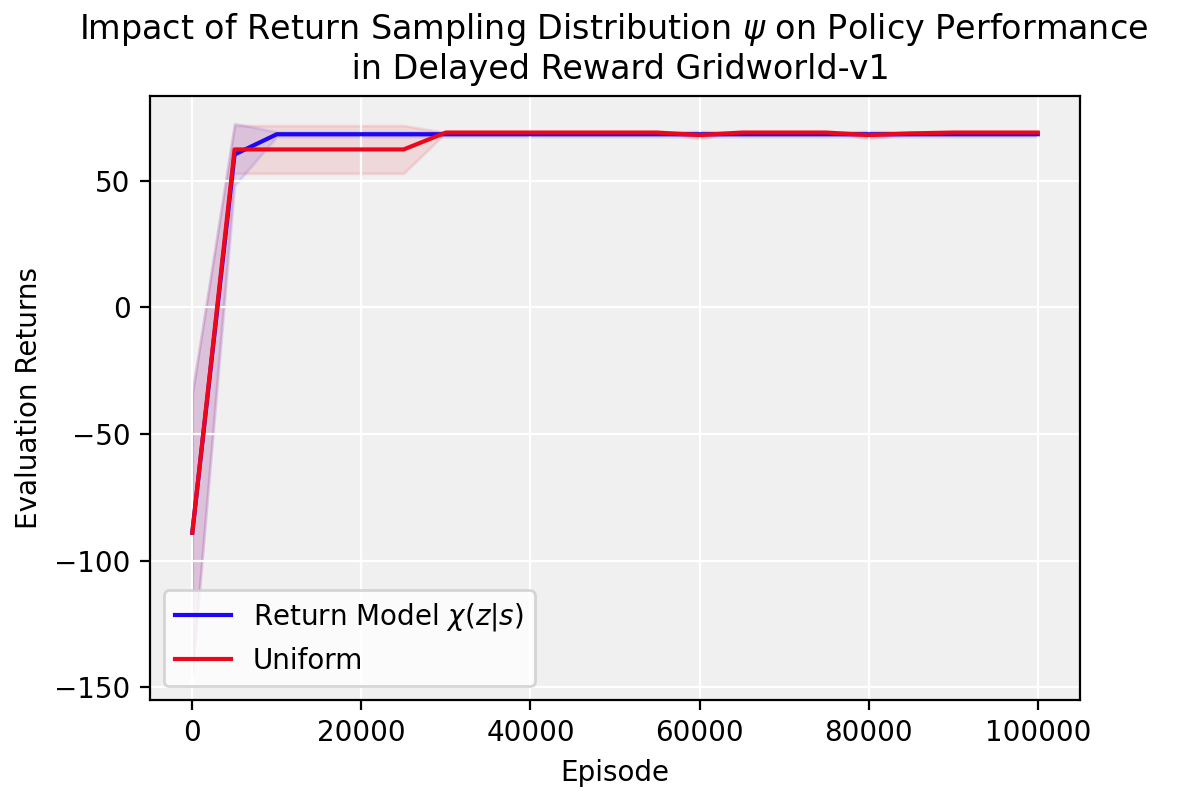}
  % \caption{A subfigure}
  \label{fig:psi_sub1}
\end{subfigure}%
\begin{subfigure}{0.33\textwidth}
  \centering
  \includegraphics[width=0.9\linewidth]{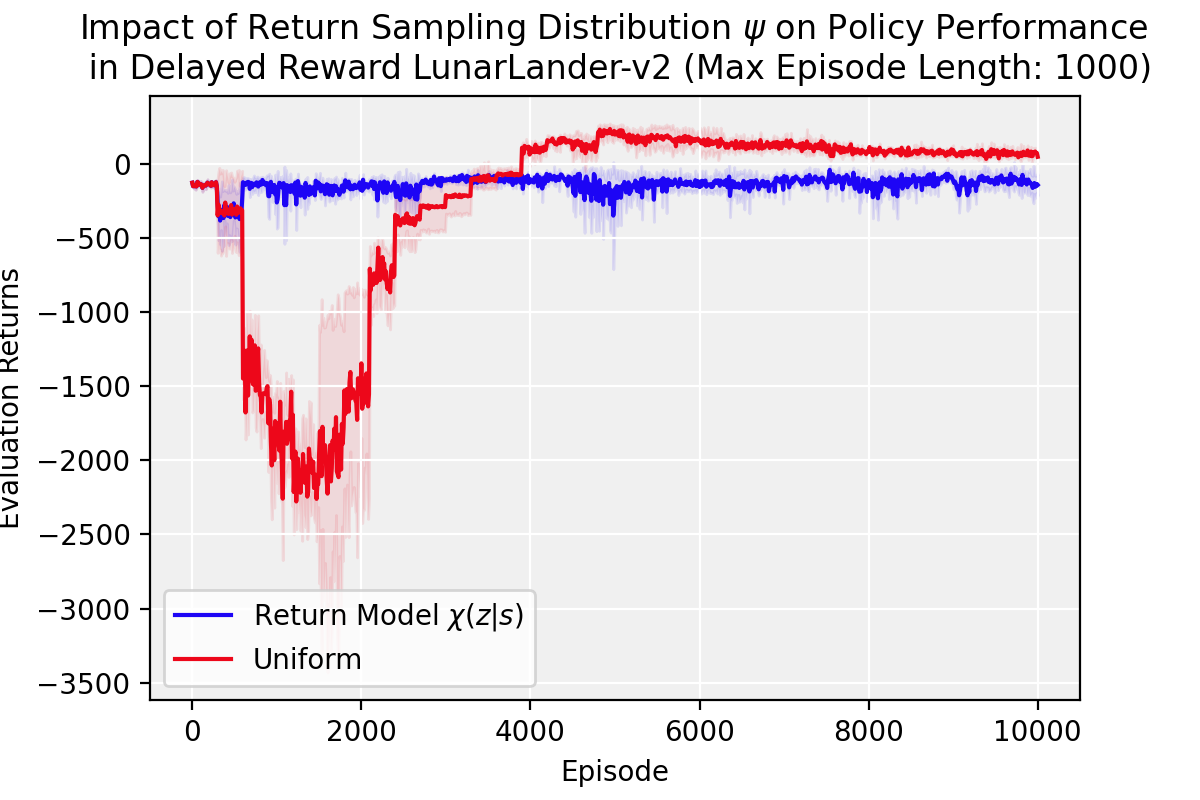}
  % \caption{A subfigure}
  \label{fig:sub2}
\end{subfigure}
\begin{subfigure}{0.33\textwidth}
  \centering
  \includegraphics[width=0.9\linewidth]{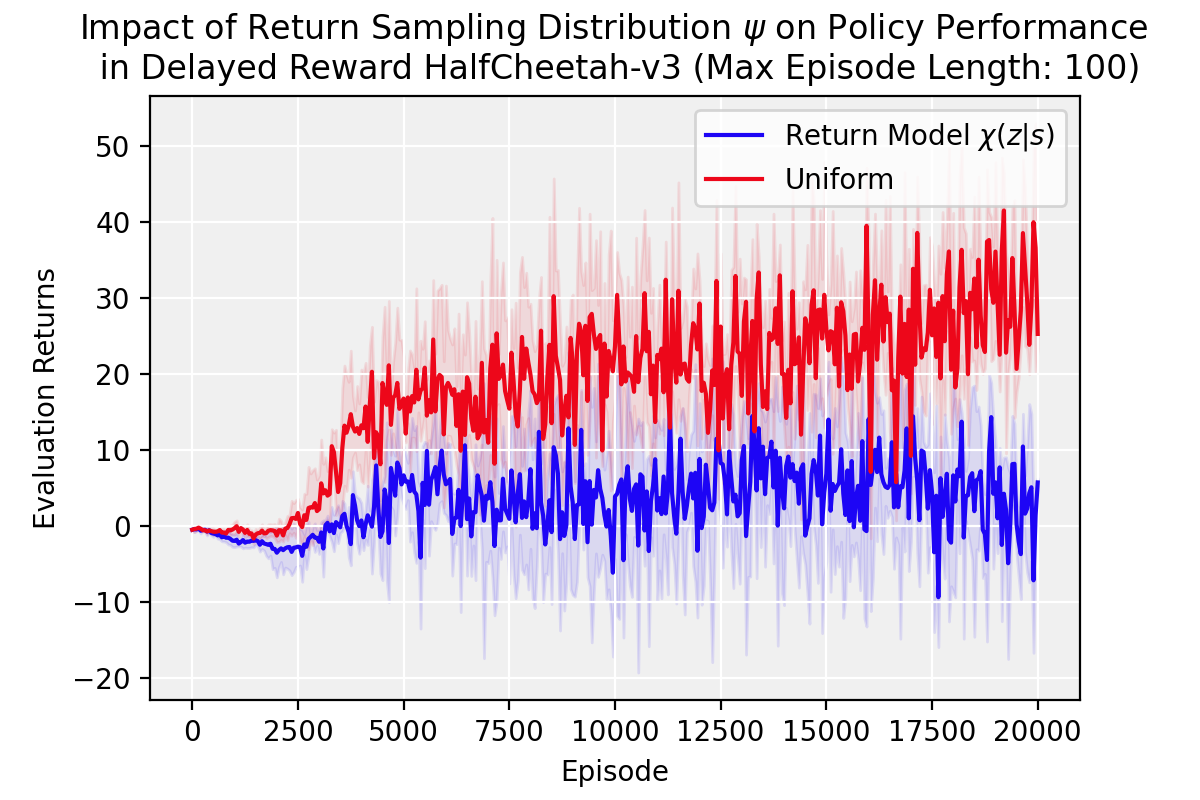}
  % \caption{A subfigure}
  \label{fig:psi_sub2}
\end{subfigure}
% \vspace{-15pt}
\caption{Impact of return sampling distribution $\psi$ on policy performance in Gridworld-v1 and delayed reward LunarLander-v2 and Half-Cheetah-v3. Sampling rewards with a uniform distribution outperforms sampling with respect to the conditional distribution $\chi(z|s)$ in the more difficult LunarLander and HalfCheetah settings.}
% \vspace{-10pt}
\label{fig:Psi-ablation}
\end{figure*}

Figure \ref{fig:Psi-ablation} examines the impact of $\psi$ -- the distribution from which returns are sampled in the second expectation in expression \ref{eq:h-dice-opt} -- on final performance in Gridworld-v1, LunarLander-v2 with delayed rewards, and Half-Cheetah-v3 with delayed rewards. All policies again are trained with H-DICE, with only the choice of $\psi$ changing between experiments. As justified in Appendix \ref{app:math}, we study two options for $\psi$: a uniform distribution over returns, and the state-conditioned return distribution $\chi^{\pi_\theta}(z|s)$. Note that computing the hindsight ratio  using the uniform distribution requires $\chi^{\pi_\theta}(z|s)$ to be multiplied with $\phi^*(s, a, z)$, while using $\chi^{\pi_\theta}(z|s)$ enables the hindsight ratio to be directly obtained from $\phi^*(s, a, z)$. Despite both choices being theoretically justified, the results in Figure \ref{fig:Psi-ablation} clearly demonstrate that sampling with the uniform distribution results results in superior performance as compared to sampling from $\chi^{\pi_\theta}(z|s)$.
We hypothesise that this is because the Hindsight DICE model alone is not sufficient to capture the complex credit assignment structures in an environment and the scaling offered by the return model is crucial in learning more expressive hindsight ratios.
This is further supported by the fact that both choices of $\psi$ perform similarly in the GridWorld but using the state-conditioned return distributions struggles in more complex environments like LunarLander and HalfCheetah.

\section{Using \methodname{} in dense reward settings}
\label{app:dense}
\begin{figure}[h!]
\centering
\begin{subfigure}{0.45\textwidth}
  \centering
  \includegraphics[width=0.9\linewidth]{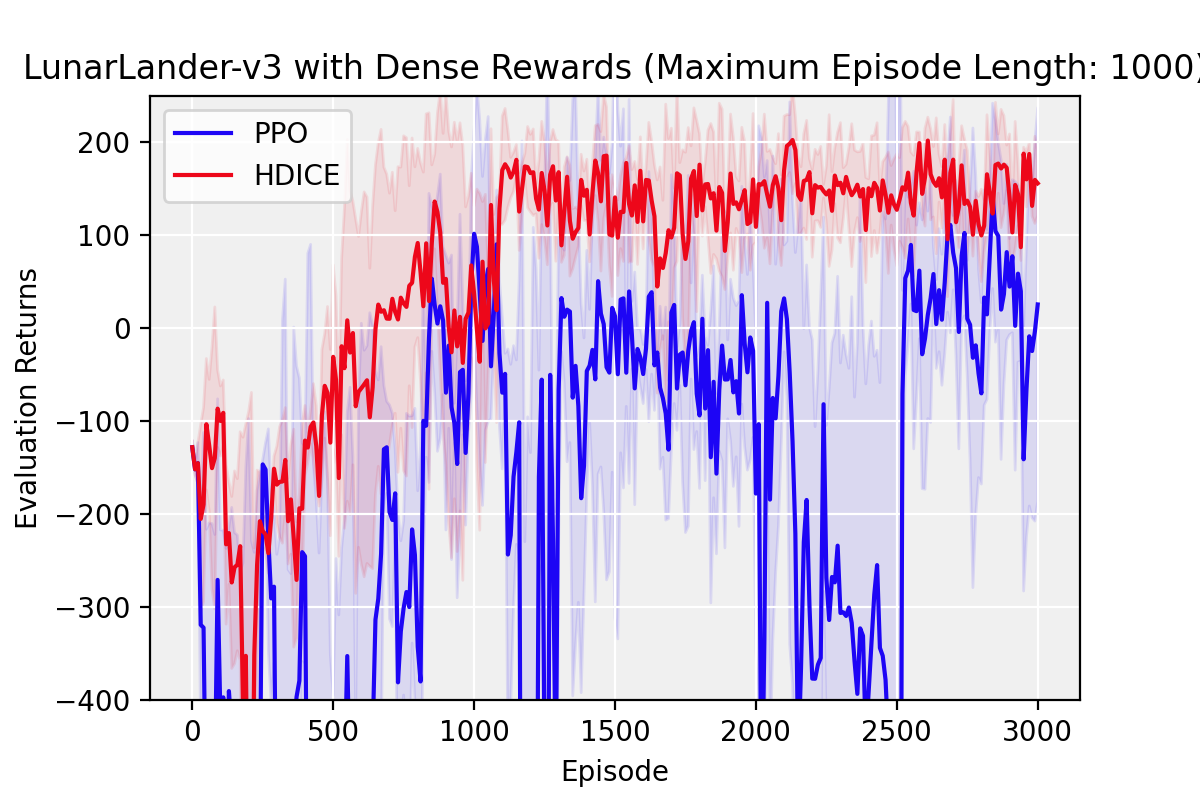}
  % \caption{A subfigure}
  \label{fig:dense_sub1}
\end{subfigure}%
\begin{subfigure}{0.45\textwidth}
  \centering
  \includegraphics[width=0.9\linewidth]{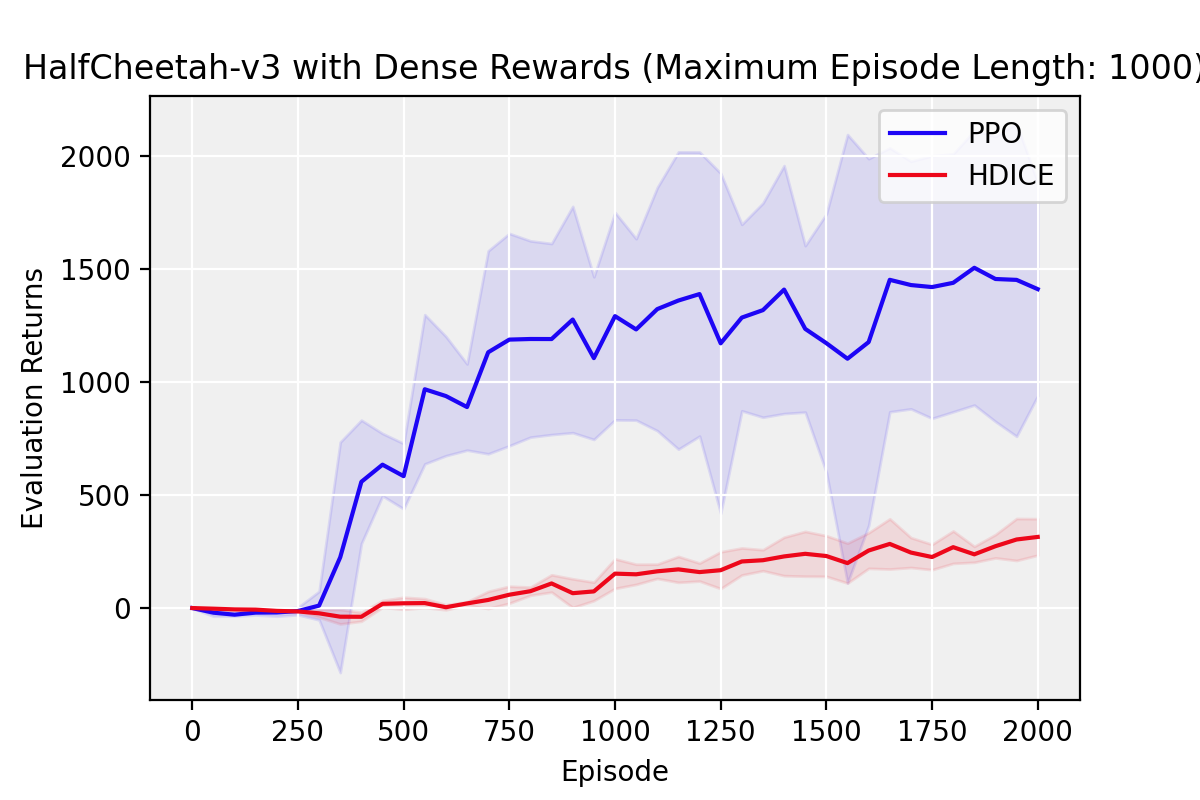}
  % \caption{A subfigure}
  \label{fig:dense_sub2}
\end{subfigure}
% \vspace{-15pt}
\caption{Performance of \methodname{} and PPO in LunarLander-v3 and HalfCheetah-v3 (both with maximum episode lengths of 1000) in dense reward settings. Note that these are the default settings of the environments. \methodname{} learns stably and quickly in LunarLander, outperforming PPO. However, \methodname{} despite learning stably, is much slower than PPO in HalfCheetah.}
\label{fig:dense-Ablation}
\end{figure}

In Figure \ref{fig:dense-Ablation}, we compare the performance of \methodname{} with PPO in a dense reward setting, where we don't anticipate a pressing need for effective credit assignment.
We seek to answer whether \methodname{} can be a competitive substitute to the GAE advantage calculation often used with PPO in a wide range of settings.
From the results we can see that \methodname{} learns more stably and achieves marginally better performance than PPO on LunarLander over three seeds but is outperformed by PPO on HalfCheetah.
We also note that \methodname{} was more stable, in terms of variation across seeds, than PPO in both environments.

\newpage

\section{Training Details and Hyperparameters}
\label{app:training}
In this section, we provide training details and hyperparameters for all environments and all methods.

To aid in the stability of training, $\chi_{\eta}^{\pi_{\theta}}, h_{\omega}^{\pi_{\theta}}$, and $\phi_{\nu}$ utilize techniques such input normalization and clipping of gradient norms. We additionally limit the output range of $\phi_{\nu}$ by applying a sigmoid at the output to improve training stability (a technique also used by \cite{nachum2019dualdice}), and also find that collecting large amounts of data between updates improved performance.  

For training the policy model, we used the usual PPO loss combined with the advantage calculation from \ref{eq:hca_advantage}. To train the Hindsight DICE model, we used the loss from equation \ref{eq:h-dice-opt}. To train the hindsight model, we used a cross-entropy loss or a Gaussian log-likelihood loss depending on whether the actions were discrete or continuous.
For the return model, we used a Gaussian log-likelihood loss to predict the mean returns from that state and used this distribution to obtain the probability of the given returns.

An extensive list of hyperparameters and the values used for each domain are provided below. While we swept across some of the key hyperparameters such as value loss coefficient, entropy coefficent, number of epochs, etc., other hyperparameters were chosen from Stable Baselines \cite{stable-baselines}. 
This is by no means an extensive hyperparameter sweep, especially for the HCA and \methodname{} models which have additional hyperparameters as compared to vanilla PPO.

We used PyTorch for our experiments and training was done on single NVIDIA RTX A5000 GPU. The code will be released on acceptance.

\subsection{GridWorld experiments}

\begin{table}[H]
\caption{Hyperparameters for the GridWorld experiments for all methods}
\centering
\begin{tabular}{|l|l|l|l|l|}
\hline
                                 & \textbf{PPO} & \textbf{PPO-HCA} & \textbf{PPO-HCA-Clip} & \textbf{\methodname{}} \\ \hline
Entropy Coefficient              & 0.1          & 0.1              & 0.1                   & 0.1             \\ \hline
Value Loss Coefficient           & 1e-4         & -                & -                     & -               \\ \hline
PPO Learning Rate                & 3e-4         & 3e-4             & 3e-4                  & 3e-4            \\ \hline
Update Every Episodes            & 50           & 50               & 50                    & 50              \\ \hline
Epsilon Clip                     & 0.2          & 0.2              & 0.2                   & 0.2             \\ \hline
PPO Epochs                       & 30           & 30               & 30                    & 30              \\ \hline
GAE Lambda                       & 0.95         & -                & -                     & -               \\ \hline
Gamma                            & 0.99         & 0.99                & 0.99                     & 0.99               \\ \hline
PPO Max Grad Norm                & -            & -                & -                     & -               \\ \hline
HCA Max Grad Norm                & -            & 10               & 10                    & 10              \\ \hline
Density Pred Max Grad Norm       & -            & -                & -                     & 10              \\ \hline
Return Pred Max Grad Norm        & -            & -                & -                     & 10              \\ \hline
Hindsight Policy Epochs                       & -            & 10               & 10                    & 10              \\ \hline
Hindsight DICE Model Epochs             & -            & -                & -                     & 10              \\ \hline
Return Model Epochs              & -            & -                & -                     & 10              \\ \hline
Auxiliary Models Learning Rate    & -            & 3e-4             & 3e-4                  & 3e-4            \\ \hline
Auxiliary Models Batchsize         & -            & 256              & 256                   & 256             \\ \hline
Return Model Normalize Targets   & -            & -                & -                     & True            \\ \hline
\end{tabular}
\label{tab:gw_hp}
\end{table}

The PPO policy was initialized with a trunk of 2 hidden layers with 64 ReLU activated hidden units each.
The value head and actor head were initialized as separate heads on top of this for PPO.
There is no value head for the HCA and \methodname{} models.

For the HCA and \methodname{} models, each auxiliary model was initialized with 2 layers with 128 ReLU activated hidden units.
In the table above, ``auxiliary models'' refers to the Hindsight Policy, Hindsight DICE Model and Return predictor together.

\subsection{LunarLander experiments}
The hyperparameters that changed from Table \ref{tab:gw_hp} are presented below. The rest of the hyperparameters are the same.

\begin{table}[H]
\caption{Hyperparameters for the LunarLander-v2 experiments for all methods}
\centering
\begin{tabular}{|l|l|l|l|l|}
\hline
                       & \textbf{PPO} & \textbf{PPO-HCA} & \textbf{PPO-HCA-Clip} & \textbf{H-DICE} \\ \hline
Entropy Coefficient    & 0.0          & 0.0              & 0.0                   & 0.01            \\ \hline
Value Loss Coefficient & 0.5          & -                & -                     & -               \\ \hline
PPO Learning Rate      & 3e-4         & 3e-5             & 3e-4                  & 3e-4            \\ \hline
Update Every Episodes  & 300          & 300              & 300                   & 300             \\ \hline
PPO Epochs             & 80           & 80               & 80                    & 80              \\ \hline
PPO Max Grad Norm      & 0.5          & 0.5              & 0.5                   & 0.5             \\ \hline
Hindsight Epochs             & -            & 20               & 20                    & 20              \\ \hline
Hindsight DICE Model Epochs   & -            & -                & -                     & 1               \\ \hline
Return Model Epochs    & -            & -                & -                     & 20              \\ \hline
\end{tabular}
\label{tab:ll_hp}
\end{table}

This time, the PPO policy was initialized with a trunk of 3 hidden layers with 128 ReLU activated hidden units each.

For the HCA and \methodname{} models, each auxiliary model was initialized with 2 layers with 128 ReLU activated hidden units.

\subsection{HalfCheetah experiments}
Once again, the hyperparameters that changes from Table \ref{tab:gw_hp} are presented below and the other hyperparameters stay the same.

\begin{table}[H]
\caption{Hyperparameters for the HalfCheetah-v3 experiments for all methods}
\centering
\begin{tabular}{|l|l|l|l|l|}
\hline
                       & \textbf{PPO} & \textbf{PPO-HCA} & \textbf{PPO-HCA-Clip} & \textbf{H-DICE} \\ \hline
Entropy Coefficient    & 0.01         & 0.01             & 0.01                  & 0.01            \\ \hline
Value Loss Coefficient & 0.5          & -                & -                     & -               \\ \hline
PPO Learning Rate      & 3e-4         & 3e-5             & 3e-4                  & 3e-4            \\ \hline
Update Every Env Steps & 6144         & 6144             & 6144                  & 6144            \\ \hline
PPO Epochs             & 80           & 80               & 80                    & 80              \\ \hline
PPO Max Grad Norm      & 0.5          & 0.5              & 0.5                   & 0.5             \\ \hline
\end{tabular}
\end{table}

The PPO policy was initialized with a trunk of 3 hidden layers with 128 ReLU activated hidden units each.

For the HCA and \methodname{} models, each auxiliary model was initialized with 2 layers with 128 ReLU activated hidden units.

\newpage
\section{GridWorld-v2}
We provide an image of GridWorld-v2 in this section. Note that the dynamics and reward structure of GridWorld-v2 are the same as in GridWorld-v1 as described in Section \ref{Sec: domains} with the exception of the maximum episode length, which is set to 100 in GridWorld-v2 (as opposed to 50 in GridWorld-v1).

\begin{figure}[h!]
    \centering
    \includegraphics[width=1.0\textwidth]{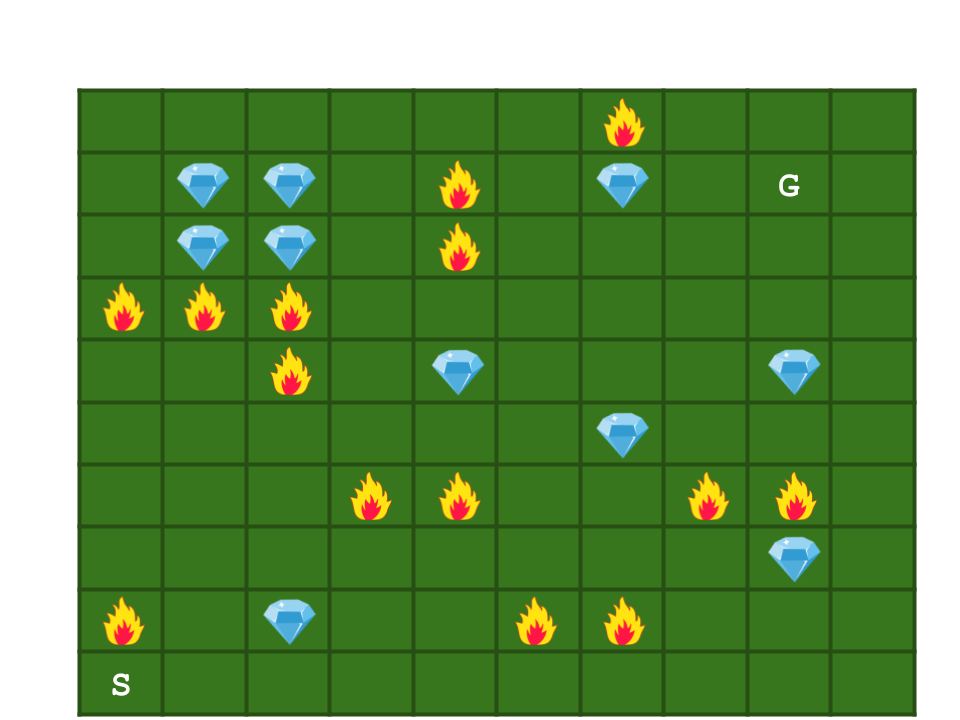} \hspace{2pt}
    \caption{The GridWorld-v2 Environment} 
    \label{fig:Gridworld-v2-img}
\end{figure}

\end{document}